\documentclass[10pt,twocolumn,letterpaper]{article}

\usepackage[pagenumbers]{cvpr} 
\usepackage[accsupp]{axessibility}  

\definecolor{cvprblue}{rgb}{0.21,0.49,0.74}
\usepackage[pagebackref,breaklinks,colorlinks,allcolors=cvprblue]{hyperref}
\usepackage{multirow}
\usepackage{tikz}
\usepackage{amsmath}

\def\paperID{1388
} 
\def\confName{CVPR}
\def\confYear{2025}

\title{FG$^2$: Fine-Grained Cross-View Localization by Fine-Grained Feature Matching}

\author{Zimin Xia \ \ \ \ \ 
Alexandre Alahi \\
École Polytechnique Fédérale de Lausanne (EPFL), Switzerland\\
{\tt\small \{zimin.xia,alexandre.alahi\}@epfl.ch}
}

\begin{document}
\maketitle
\begin{abstract}
We propose a novel fine-grained cross-view localization method that estimates the 3 Degrees of Freedom pose of a ground-level image in an aerial image of the surroundings by matching fine-grained features between the two images.
The pose is estimated by aligning a point plane generated from the ground image with a point plane sampled from the aerial image.
To generate the ground points, we first map ground image features to a 3D point cloud. 
Our method then learns to select features along the height dimension to pool the 3D points to a Bird's-Eye-View (BEV) plane. 
This selection enables us to trace which feature in the ground image contributes to the BEV representation.
Next, we sample a set of sparse matches from computed point correspondences between the two point planes and compute their relative pose using Procrustes alignment. 
Compared to the previous state-of-the-art, our method reduces the mean localization error by 28\% on the VIGOR cross-area test set.
Qualitative results show that our method learns semantically consistent matches across ground and aerial views through weakly supervised learning from the camera pose.

\end{abstract}    
\section{Introduction}
\label{sec:intro}

Visual localization aims to estimate the camera pose of a given image relative to a representation of a scene. 
For large-scale applications like autonomous driving and outdoor mobile robotics, the scalability of this representation is crucial, as mapping is expensive~\cite{zang2024data}.
Recently, using aerial imagery for accurate visual localization~\cite{shi2022accurate,xia2022visual,wang2023view,shi2023boosting,lentsch2023slicematch,xia2023convolutional,song2024learning,wang2024fine,sarlin2024snap,wang2024view,xia2024adapting,shi2024weakly} has gained increasing interest due to its global coverage.
This task, named fine-grained cross-view localization, aims to estimate the 3 Degrees of Freedom (DoF) pose, i.e., the planar location and yaw orientation, of a query ground-level image on an aerial image that covers the ground camera's local surroundings.
In practice, the aerial image can be identified using noisy position estimates from onboard Global Navigation Satellite System (GNSS) in vehicles or robots, which contain errors up to tens of meters in urban canyons~\cite{benmoshe2011urbangnss}.

In fact, fine-grained cross-view localization is common in everyday life, such as when people use Google Satellite Maps~\cite{googlemaps} to determine their position. 
In this process, people first identify surrounding landmarks, such as buildings and lane markings, understand their relative positions, and then compare them to the aerial image. However, existing methods do not fully follow this intuitive approach.

Global descriptor-based methods~\cite{xia2022visual,xia2023convolutional,lentsch2023slicematch} generate descriptors at candidate locations in the aerial image and compare them to the descriptor of the query ground image. 
Intuitively, this approach mimics imagining what would be visible in the ground view at each candidate location given the aerial view but is not how humans naturally approach localization. 
Local feature-based methods typically transform the ground image into a Bird's-Eye-View (BEV) representation using homography~\cite{wang2023view,song2024learning}. 
However, homography applies only to the ground plane, limiting these methods by overlooking above-ground objects, such as buildings.
Besides, some methods~\cite{fervers2023uncertainty,shi2023boosting} estimate the ground camera's pose by densely comparing the ground view BEV representation to the aerial view. 
Yet humans tend to localize based on a few distinct objects~\cite{wolbers2010determines,epstein2014neural}, rather than the entire panoramic scene. 
Moreover, no existing method explicitly identifies which objects in the ground view match those in the aerial view, making localization results hard to interpret.

\begin{figure}
    \centering
    \includegraphics[width=1\linewidth]{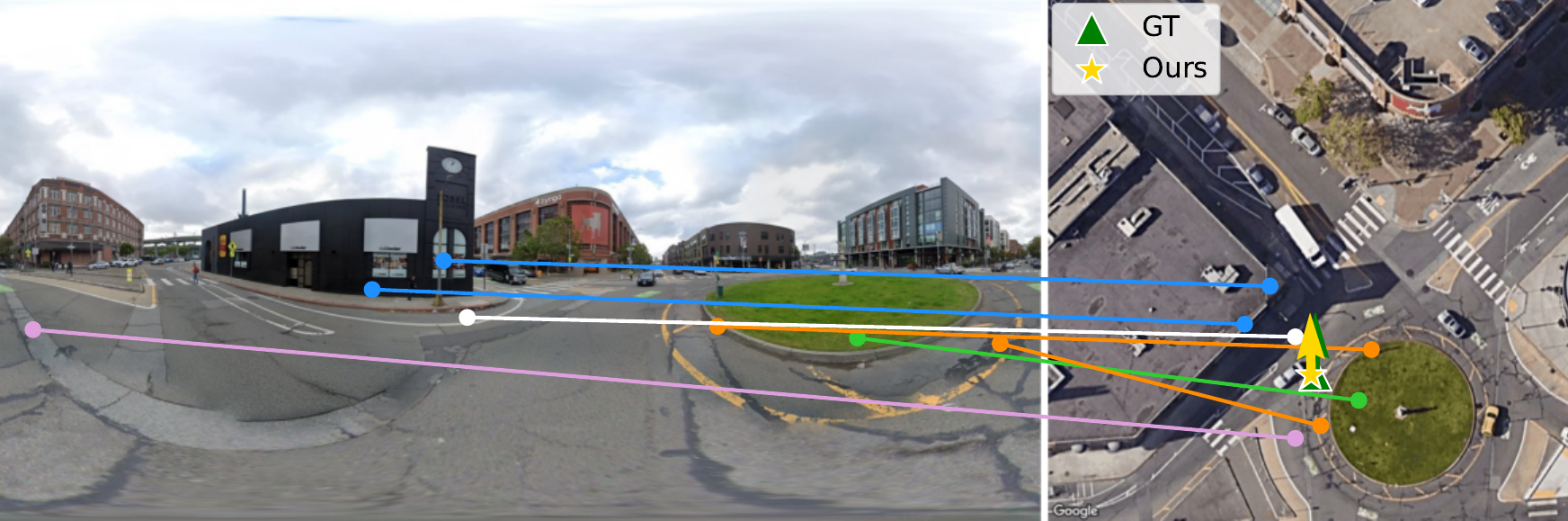}
    \captionof{figure}{Fine-grained cross-view localization estimates the 3 DoF pose of a ground image (left) on an aerial image (right) of the surroundings. Our method tackles this task by matching fine-grained local features across views, providing interpretable results. The matched correspondences are semantically consistent and learned \textit{without ground truth labels}. Here are a few selected predictions.
    }
    \label{fig:figure1}
\end{figure}

We bridge the above gaps by proposing a novel fine-grained cross-view localization method that aligns with human intuition. 
As shown in Fig.~\ref{fig:figure1}, our method matches fine-grained local features across views and solves localization through sparse correspondences.
Similar to prior works, we also transform the ground-level image into a BEV representation.
However, our method is unique such that it first explicitly generates a 3D point cloud centered at the ground camera and then learns to select features along the height dimension, pooling this 3D point cloud into a point plane. 
This enables us to trace which features in the ground-level image contribute to the point plane.
Next, our method computes the correspondences between the generated points from the ground view and points sampled from the aerial view and solves the relative pose between the two point planes using Procrustes alignment~\cite{schonemann1966generalized} with sparse correspondences.
Notably, without any ground truth local feature correspondences between ground and aerial images, our method learns reasonable matches in a weakly supervised manner using only the ground truth camera pose.

Concretely, our main contributions are\footnote{Code is available at: \url{https://github.com/vita-epfl/FG2}}:
(i) A novel fine-grained cross-view localization method that matches local features between ground and aerial images. Experiments show that our method achieves superior localization accuracy, reducing the mean localization error by 28\% on a challenging benchmark compared to the previous state-of-the-art.
(ii) Our proposed feature selection along height enables the interpretability of the localization results. With supervision only on camera pose, our method learns to match local features between ground and aerial images in a weakly supervised manner.
Qualitative results show that our matching is semantically consistent across ground and aerial views.

\section{Related work}
\label{sec:related_work}

\textbf{Deep cross-view localization methods}
typically start with a Siamese-like feature extractor.
Built on top of the extracted features, methods have two different objectives for constructing representations to localize the ground images.

\textit{Representing what should be seen} by a ground camera positioned at candidate locations in an aerial image is a common formulation~\cite{xia2022visual,lentsch2023slicematch,xia2023convolutional}.
These methods render descriptors at candidate locations in the aerial view and then match them with the descriptor of the ground view to generate a localization heat map.
During this step, \cite{lentsch2023slicematch,xia2023convolutional} constructs orientation-aware descriptors to estimate the camera's orientation jointly.
The idea of learning a global descriptor to summarize expected visual content is also used in other BEV map-based localization~\cite{howard2021lalaloc,min2022laser,howard2022lalaloc++} and cross-view image retrieval~\cite{hu2018cvm,shi2019spatial,xia2020geographically,yang2021cross,shi2022beyond,mi2024congeo,fervers2024statewide}.
Although the heat maps indicate localization uncertainty, interpreting which objects contribute to localization remains challenging due to the use of abstract global image descriptors.

\textit{Representing what is here} refers to constructing a BEV representation from the ground view, where each BEV cell captures details of the local vicinity~\cite{fervers2023uncertainty,wang2023view,sarlin2024snap,wang2024view,song2024learning,shi2023boosting,wang2024fine,fervers2023c}. 
Localization is then achieved in two ways. 
The first approach sweeps the ground view BEV representation across the aerial BEV representation and identifies the pose with the highest alignment~\cite{fervers2023uncertainty,sarlin2024snap,fervers2023c,shi2023boosting}. 
However, supervising element-wise alignment between two BEV representations encourages the model's ground branch to hallucinate information behind occlusions.
The second approach estimates the relative pose between ground and aerial BEV features using dense~\cite{song2024learning,wang2024fine} or sparse~\cite{wang2023view,wang2024view} matches between them. 
However, \cite{wang2023view,song2024learning} apply homography in BEV mapping, forcing the model to ignore above-ground objects. 
\cite{wang2024view} use transformers to resolve ambiguity along height, but it assumes knowledge of on-ground pixels in the ground view.
In practice, the height of horizon in the image depends on occlusions along the depth, so this assumption does not hold.
Moreover, \cite{wang2024fine,song2024learning} use learned modules to predict the pose, making the results hard to interpret, as it is unclear how specific features contribute to the prediction.

\textbf{Ground-level image relative pose estimation} is a simpler yet closely related task. 
The classic pipeline~\cite{hartley2003multiple} involves detecting and describing key points in two images, matching these 2D key points, and estimating the 6 DoF pose based on the matches within RANSAC loops~\cite{fischler1981random}. 
Recent advancements~\cite{leroy2024grounding,barroso2024matching,wang2024dust3r} lift the two images to two sets of 3D points and align the two 3D point clouds for relative pose estimation.
For ground-to-aerial cross-view localization (which focuses on 3 DoF pose), one can generate two sets of points on a BEV plane and then align two point planes based on their sparse point-wise matches.
\section{Methodology}

Given a ground-level image $G$ and an aerial image $A$ that covers the ground camera's local surroundings, the objective of fine-grained cross-view localization is to estimate the 3 DoF pose $\boldsymbol{p}=[\boldsymbol{t}, o]$ of the ground camera, where $\boldsymbol{t}$ represents the pixel coordinates of camera's location in the aerial image $A$, and $o$ denotes the camera's yaw orientation.

\begin{figure*}[t]
    \centering
    \includegraphics[width=1\linewidth]{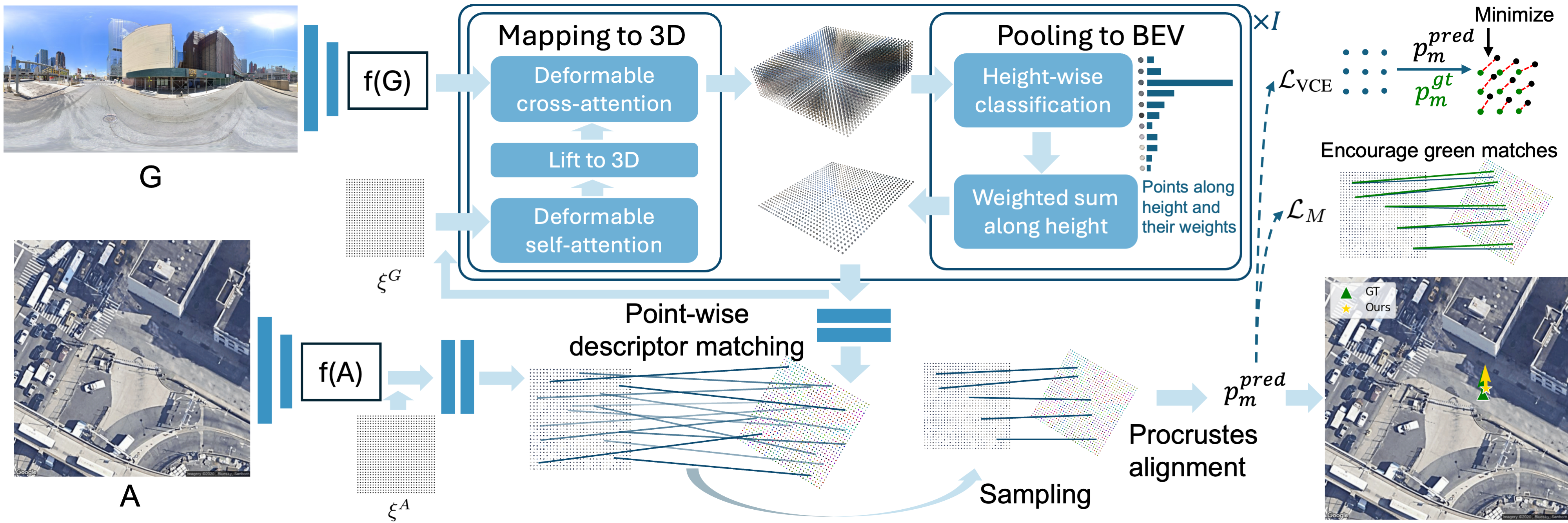}
    \caption{Overview of our proposed method and the objectives of the loss functions used. We define two sets of points, $\xi^G$ and $\xi^A$, on a BEV plane, and our method generates a descriptor for each point. For the ground view, this involves mapping the image feature $f(G)$ to 3D and then selecting the important features along height. Next, we compute pairwise matching scores between the two point sets. The pose is then computed using Procrustes alignment based on the sampled matches. $\mathcal{L}_{\text{VCE}}$ minimizes the difference between a virtual point set transformed using the predicted pose and the ground truth pose. $\mathcal{L}_{M}$ encourages correspondencs found using the ground truth pose.}
    \label{fig:method}
\end{figure*}

\subsection{Overview}
\textbf{Motivation:}
Our goal is to achieve fine-grained cross-view localization by matching fine-grained features across views. 
However, this matching has two main challenges: 
(i)~Each pixel in the aerial image corresponds to a vertical ray in 3D (assuming parallel projection).
The matching method needs to identify the height along this vertical line observed in the ground view that matches aerial view content. 
However, existing datasets do not have such ground truth correspondences; 
(ii)~Matchable information is sparse since not all content in the aerial view is visible from the ground view due to occlusions.
Besides, dynamic objects are unmatchable because the images are not captured simultaneously.

We address the above challenges with a model that is end-to-end trained using only the ground truth camera pose. 
To resolve the height ambiguity, our method first lifts the ground image to a 3D point cloud and then selects the 3D points along the vertical direction whose feature best aligns with the aerial view.
This selection is learned from data such that the selected feature should produce a matching result in the next step that aligns with the ground truth camera pose.
This selection pools the 3D point cloud into a 2D point plane.
We then match these 2D points with 2D points sampled from the aerial view. 
Given that the matchable information is sparse, we select only the top correspondences to compute the ground camera pose.
Notably, our explicit selection of the vertical points allows us to trace which objects in the ground image are matched to those in the aerial image, providing interpretable localization results. 

\textbf{Formulation:}
We estimate the ground camera pose $\boldsymbol{p}=[\boldsymbol{t}, o]$ by aligning two point planes, i.e., computing their metric relative pose $\boldsymbol{p}_{m} = [\boldsymbol{t}_{m}, o]$. 
One point plane has its origin at the ground camera and the other has its origin at the center of the aerial image.
Given the known Ground Sampling Distance (GSD) in meters per pixel of the aerial image, we compute $\boldsymbol{t}$ as $\boldsymbol{t} = \boldsymbol{t}_{m} / \text{GSD} + \boldsymbol{s}$, where $\boldsymbol{s}$ is the offset between the center and the top-left of the aerial image.

As shown in Fig.~\ref{fig:method}, our method first extracts ground and aerial image features $f(G)$ and $f(A)$.
Similar to recent advances in ground-level image matching~\cite{barroso2024matching,edstedt2024dedode}, we adopt pre-trained DINOv2 features\footnote{Our Appendix will show that directly matching ground and aerial DINO features without our designed BEV points matching does not work.}~\cite{oquab2023dinov2} as $f(G)$ and $f(A)$.
To construct the ground and aerial point planes, we define two sets of $N$ points, $\xi^G=\{(x^G_1, y^G_1), ..., (x^G_N, y^G_N)\}$ and $\xi^A=\{(x^A_1, y^A_1), ..., (x^A_N, y^A_N)\}$, regularly spanning an area $[-L/2, L/2]^2$, and $L$ is the side length of the aerial image in meters. 
$(x, y)$ denotes the coordinates of each point, and the origins of $\xi^G$ and $\xi^A$ are located at the ground camera and the center of the aerial image, respectively.
Next, there are two main steps to infer the 3 DoF relative pose $\boldsymbol{p}_{m}$ between $\xi^G$ and $\xi^A$: 
First, we generate a descriptor for each point in $\xi^G$ and $\xi^A$ (Sec.~\ref{sec:bev_mapping}).
Second, we compute $\boldsymbol{p}_{m}$ based on the point matching scores~(Sec.~\ref{sec:pose_estimation}).

\subsection{BEV mapping}
\label{sec:bev_mapping}

Given the two point sets, $\xi^G$ and $\xi^A$, and ground and aerial image features, $f(G)$ and $f(A)$, our BEV mapping generates for each ground and aerial point, $(x^G_n, y^G_n)$ and $(x^A_n, y^A_n)$, a descriptor, $d^G_n$ and $d^A_n$, where $n\in[1,...,N]$.
For the ground view, we first map the ground image feature to 3D and then pool the 3D feature to BEV.

\textbf{Mapping to 3D:}
We follow an approach similar to recent 3D object detection~\cite{li2022bevformer,wang2024unibev} to map image features to 3D.
Each 2D point in $\xi^G$ is lifted to a pillar of $M$ 3D points spanning a given height range.
These 3D points are then projected to the ground image to retrieve features.

Specifically, we use deformable attention~\cite{zhu2021deformable}.
For the $n$-th point, $(x^G_n, y^G_n)$, in $\xi^G$, we first associate it with a learnable query $q_n^G$. 
Next, we find for the $m$-th 3D point $\mathcal{X}_{n,m}$, $m \in [1,...,M]$, its projected pixel coordinates $\boldsymbol{r} = (u_{n,m}, v_{n,m})$ in $f(G)$.
Then, $\mathcal{X}_{n,m}$ retrieves features from $f(G)$ using deformable cross-attention,
\begin{align}
    F_{\mathcal{X}_{n,m}} = \sum^J_{j=1} W_j  \sum^K_{k=1} A_{j,k} \times f(G)_{\boldsymbol{r} + {\Delta \boldsymbol{r}_{j,k}}}.
    \label{eq:deformable_atten}
\end{align}

In Eq.~\ref{eq:deformable_atten}, $F_{\mathcal{X}_{n,m}}$ is the feature for the 3D point $\mathcal{X}_{n,m}$. 
The deformable attention has $J$ attention heads with attention weights $W_j$ generated from the query $q_n^G$.
Within each head, there are $K$ offsets $\Delta \boldsymbol{r}_{j,k}$ and their weights $A_{j,k}$ mapped from the query $q_n^G$, where $\sum^K_{k=1} A_{j,k} = 1$.
The term $f(G)_{\boldsymbol{r} + {\Delta \boldsymbol{r}_{j,k}}}$ denotes the bilinearly interpolated feature at the location $\boldsymbol{r} + {\Delta \boldsymbol{r}_{j,k}}$ in $f(G)$.
These learnable offsets $\Delta \boldsymbol{r}_{j,k}$ allow the model to capture informative features around the projected pixel coordinates $\boldsymbol{r}$.

We retrieve features for all 3D points.
If the ground image $G$ is panoramic, all points can find their 2D pixels inside $f(G)$.
When $G$ has a limited field of view, we record any 3D points projected outside $f(G)$ and discard them when performing ground-aerial point matching in later pose estimation. 
Similar to~\cite{li2022bevformer}, before mapping to 3D, the learnable query $q_n^G$ first undergoes a deformable self-attention layer to gather information from neighboring 2D points, i.e., using Eq.~\ref{eq:deformable_atten} by replacing $f(G)$ with the queries of 2D points.

\textbf{Pooling to BEV:}
Next, we pool the 3D points to the BEV plane and generate a descriptor $d^G_n$ for each 2D point $(x^G_n, y^G_n)$.
Typically, this is done by summing~\cite{li2022bevformer} or taking max~\cite{sarlin2024snap} feature among the $M$ 3D points over height.
However, we argue that both approaches discard information about which height contributes to the BEV feature, as the dominant feature in summation or the argmax can vary across channels.
To identify the ground feature along height that contributes to matching with the aerial view, we propose to learn the feature selection along height.
Our ablation study will show it also improves localization accuracy.

Given the 3D points $\mathcal{X}_{n,m}$ and their features $F_{\mathcal{X}_{n,m}}$, we treat the BEV pooling problem as a classification problem.
We classify the importance of the $M$ 3D points along the height dimension and the feature $g^G_n$ for the 2D point $(x_n^G, y_n^G)$ is then a weighted sum of the $M$ features,
\begin{align}
    g^G_n = \sum^M_{m=1} \frac{e^{l(F_{\mathcal{X}_{n,m}})}}{\sum^M_{m=1}{e^{l(F_{\mathcal{X}_{n,m}})}}}  F_{\mathcal{X}_{n,m}},
\end{align}
where $l$ maps each $F_{\mathcal{X}_{n,m}}$ to a single value.
We highlight again that we do not have ground truth for this classification, and the model needs to learn the feature selection backpropagated from the supervision on the camera pose.

In practice, we treat the mapping to 3D and pooling to BEV as a unified module and iterate it $I$ times (using $g^G_n$ as the new query $q^G_n$), similar to~\cite{li2022bevformer}.
Finally, we obtain the descriptor $d^G_n$ of the 2D point $(x_n^G, y_n^G)$ by applying a projection head on its final feature, $d^G_n = \text{proj}(g^G_n)$, and the projection head consists of ResNet blocks~\cite{he2016deep}, a small self-attention layer, and a $L_2$-normalization layer.

\textbf{Aerial BEV sampling:}
Since the aerial view already provides a BEV perspective, assigning features to points in $\xi^A$ only requires looking up the feature at the corresponding locations in the extracted aerial image feature map $f(A)$. 
Specifically, for each point $(x_n^A, y_n^A)$, its feature $g^A_n$ is a bilinear interpolation at location $(\frac{(x_n^A, y_n^A)}{\text{GSD}} + \boldsymbol{s})$ in $f(A)$. 
Similar to the ground branch, we use a projection head with the same architecture to map  $g^A_n$ to a descriptor $d^A_n$.

\subsection{Pose estimation}
\label{sec:pose_estimation}

We compute the metric pose $\boldsymbol{p}_{m}$ by first matching descriptors of two point sets $\xi^G$ and $\xi^A$, and then use Procrustes alignment~\cite{schonemann1966generalized} based on matched correspondences.

\textbf{Point-wise descriptor matching:}
Given the ground and aerial point descriptors, $\{d^G_1, ..., d^G_N\}$ and $\{d^A_1, ..., d^A_N\}$, we first compute their pair-wise similarly score $c_{n_G, n_A}$, 
\begin{align}
    c_{n_G, n_A} = \frac{\text{cosine similarity}(d^G_{n_G}, d^A_{n_A})}{\tau}, 
    \label{eq:similarity}
\end{align}
where $\tau$ is a temperature parameter and $n_G \in [1,N], n_A \in [1,N]$ are indexes for ground and aerial descriptors.

Since the descriptor matching probability is the probability that they are mutual nearest neighbors, 
we use the dual-Softmax operator as in~\cite{edstedt2024dedode,sun2021loftr,tyszkiewicz2020disk,barroso2024matching} to compute the product of the probability of $d^G_{n_G}$ matched to $d^A_{n_A}$ for a fixed $n_G$ and the probability of $d^A_{n_A}$ matched to $d^G_{n_G}$ for a fixed $n_A$,
\begin{align}
    D_{n_G,n_A} = \frac{e^{c_{n_G, n_A}}}{\sum^{N}_{n_A=1}{e^{c_{n_G, n_A}}} + e^b} \times \frac{e^{c_{n_G, n_A}}}{\sum^{N}_{n_G=1}{e^{c_{n_G, n_A}}} + e^b}.
\end{align}

The term $D_{n_G,n_A}$ denotes the probability of the $n_G$-th ground point $(x_{n_G}^G, y_{n_G}^G)$ and the $n_A$-th aerial point $(x_{n_A}^A, y_{n_A}^A)$ are matched and $b$ is a learnable dustbin parameter to allow unmatched points to be assigned to it~\cite{sarlin2020superglue}.

\textbf{Pose estimation by Procrustes alignment: }
We use Kabsch solver~\cite{kabsch1976solution} (details in Appendix) to compute the 3 DoF pose $\boldsymbol{p}_{m}$ between the two point sets, $\xi^G$ and $\xi^A$, given their correspondences.
This process is differentiable~\cite{avetisyan2019end,brachmann2021visual}.

Since not all information in the ground and aerial views are matchable, we sample a subset of $N_S$ correspondences for Procrustes based on the descriptor matching probability $D$.
Our ablation study will show that sampling a subset provides better localization accuracy than using all points.

\textbf{Inference with RANSAC:}
As our pose estimation requires sampling from the matching probability $D$, we can incorporate robust estimators, such as RANSAC~\cite{fischler1981random}, in inference.
We sample the descriptor matching probability $R$ times.
In each iteration $r$, we count the inliers for the pose $\boldsymbol{p}_{m}^{r}$, computed using the sampled correspondences.
A ground-aerial point correspondence $(x^G_{n_G}, y^G_{n_G}) \leftrightarrow (x^A_{n_A}, y^A_{n_A})$ is considered as an inlier if,
\begin{align}
    \sqrt{(x^A_{\tilde{n_G}} - x^A_{n_A})^2 + (y^A_{\tilde{n_G}} - y^A_{n_A})^2} < \text{threshold},
    \label{eq:inlier}
\end{align}
where $(x^A_{\tilde{n_G}}, y^A_{\tilde{n_G}})$ is acquired by transforming the ground point $(x^G_{n_G}, y^G_{n_G})$ using $\boldsymbol{p}_{m}^{r}$.
Finally, we find the pose with the most inliers and use all inliers to compute $\boldsymbol{p}_{m}^{pred}$.

\subsection{Loss}
Our loss combines a Virtual Correspondence Error loss adapted from~\cite{arnold2022map} and a matching loss using a weight $\beta$,
\begin{align}
    \mathcal{L} = \mathcal{L}_{\text{VCE}} + \beta\mathcal{L}_{M}.
\end{align}

Our $\mathcal{L}_{\text{VCE}}$ provides supervision signal on the predicted pose $\boldsymbol{p}_{m}$.
As shown in Fig.~\ref{fig:method} right, we define a set of 2D points (in blue) spanning a BEV space $[-L_{V}/2, L_{V}/2]^2$, where $L_{V}$ is a hyperparameter. 
Then, we use the ground truth $\boldsymbol{p}^{gt}_{m}$ and predicted pose $\boldsymbol{p}^{pred}_{m}$ to find the true (in green) and predicted (in black) correspondences for the blue points.
Our $\mathcal{L}_{\text{VCE}}$ then minimizes the mean Euclidean distance between all corresponding green and black points.

Besides supervising the camera pose, we use $\mathcal{L}_{M}$ to provide a direct supervision signal on the ground-aerial point descriptor matching.
As the matchable information between ground and aerial images is sparse, not all points in $\xi^G$ should find a correspondence in $\xi^A$ and vice versa.
Therefore, we take the $N_S$ sampled correspondences to compute $\mathcal{L}_{M}$.
For each estimated correspondence $(x^G_{n_G}, y^G_{n_G}) \leftrightarrow (x^A_{n_A}, y^A_{n_A})$, we find for the ground point $(x^G_{n_G}, y^G_{n_G})$ the ground truth aerial point $(x^A_{\hat{n_G}}, y^A_{\hat{n_G}})$ using $\boldsymbol{p}^{gt}_{m}$ and then compute an infoNCE loss~\cite{oord2018representation} for all $N_S$ correspondences,
\begin{align}
    \mathcal{L}^G_{\text{infoNCE}} = -\log \frac{\sum_{n_G=1}^{N_S} e^{c_{n_G, \hat{n_G}}}}{\sum_{n_G=1}^{N_S} \sum_{n_A=1}^{N}{e^{c_{n_G, n_A}}}},
\end{align}
where $c$ is the descriptor similarity score~(Eq.~\ref{eq:similarity}). 
Similarly, we compute an infoNCE loss by finding the ground truth ground point $(x^G_{\hat{n_A}}, y^G_{\hat{n_A}})$ for each aerial point $(x^A_{n_A}, y^A_{n_A})$, 
\begin{align}
    \mathcal{L}^A_{\text{infoNCE}} = -\log \frac{\sum_{n_A=1}^{N_S} e^{c_{n_A, \hat{n_A}}}}{\sum_{n_A=1}^{N_S} \sum_{n_G=1}^{N}{e^{c_{n_G, n_A}}}}.
\end{align}

Finally, $\mathcal{L}_{M}$ is an average of the two infoNCE losses,
\begin{align}
    \mathcal{L}_{M} =  (\mathcal{L}_{\text{infoNCE}}^G + \mathcal{L}_{\text{infoNCE}}^A ) / 2.
\end{align}
\section{Experiments}
We first introduce the used datasets and evaluation metrics. 
Then, our implementation details are presented. 
Next, we compare our method to previous state-of-the-art and show ground-aerial fine-grained feature matching results. 
Finally, we conduct an ablation study on key hyperparameters.

\subsection{Datasets and evaluation metrics}
We use two datasets in our experiments.

\textbf{VIGOR}~\cite{zhu2021vigor} contains ground-level panoramas and aerial images collected in four cities in the US. 
It has two settings: same-area and cross-area.
In the same-area setting, the training and test images come from all cities, whereas, in the cross-area setting, training images are from two cities, and test images are from the other two.
Consistent with previous evaluation protocol~\cite{xia2023convolutional,lentsch2023slicematch}: we include both known and unknown orientation settings; use the positive samples, i.e., ground images located in the center $1/4$ region in the aerial images; split $20\%$ of training data for validation and ablation study; and use the ground truth labels from~\cite{lentsch2023slicematch}.

\textbf{KITTI}~\cite{Geiger2013IJRR} provides front-facing images with a limited field of view collected in Germany.
\cite{shi2022accurate} provides aerial images for KITTI.
Similar to VIGOR~\cite{zhu2021vigor}, the data is split into same-area and cross-area.
We adopt the common setting~\cite{shi2022beyond} that ground images are located in the center $40~\text{m} \times 40~\text{m}$ area of their corresponding aerial images, and there is an orientation prior with noise between $\pm10^\circ$.

\textbf{Metrics:}
We report mean and median localization and orientation errors. Following~\cite{xia2023convolutional, wang2024fine}, on KITTI, we further decompose localization errors into longitudinal and lateral components based on the driving direction, and report the percentage of samples within given error thresholds.

\subsection{Implementation details}
On VIGOR, we have $41 \times 41$ points in both $\xi^G$ and $\xi^A$, and each ground 2D point is lifted to $11$ 3D points, regularly distributed from $-20$~m to $20$~m (assuming the camera is at $0$~m).
On KITTI, we use the same number of points in $\xi^A$. 
Since KITTI's ground image has a limited field of view, we create ground points only in front of the camera, resulting in $21 \times 41$ points in $\xi^G$.
The $11$ 3D points span from $-25$~m to $25$~m to ensure all content in the image can be mapped to 3D. 
Similar to~\cite{li2022bevformer}, we iterative the unified BEV mapping module $I=6$ times.
As in~\cite{barroso2024matching}, we use $\tau = 10$ in Eq.~\ref{eq:similarity}.
We sample $N_S = 1024$ correspondences and keep their matching scores as weights when solving Procrustes alignment.

On both datasets, we use AdamW~\cite{loshchilov2017decoupled} with a learning rate of $1 \times 10^{-4}$, and our batch size is $24$.
The weight $\beta$ in our loss is set to $1$, and we use $10 \times 10$ 2D points over $[-2.5~\text{m}, 2.5~\text{m}]^2$ space ($L_{V}=5$ m) when computing $\mathcal{L}_{\text{VCE}}$.

\subsection{Quantitative results}

\begin{table*}[h]
    \centering
    \begin{tabular}{p{1.4cm}p{2.25cm}p{1.2cm}p{1.2cm}p{1.2cm}p{1.2cm}p{1.2cm}p{1.2cm}p{1.2cm}p{1.2cm}}
    \toprule
    \multirow{3}{*}{{Orien.}} & \multirow{3}{*}{Methods} & 
    \multicolumn{4}{c}{Same-area} & \multicolumn{4}{c}{Cross-area} \\
    \cline{3-10} 
    & & \multicolumn{2}{c}{$\downarrow$ Localization (m)} & \multicolumn{2}{c}{$\downarrow$ Orientation ($^\circ$)} & \multicolumn{2}{c}{$\downarrow$ Localization (m)} & \multicolumn{2}{c}{$\downarrow$ Orientation ($^\circ$)} \\
    \cline{3-10} 
    & & Mean & Median & Mean & Median & Mean & Median & Mean & Median \\
    \hline
    \multirow{6}{*}{Known} & 
    SliceMatch~\cite{lentsch2023slicematch} & 5.18 & 2.58 & - & - & 5.53 & 2.55 & - & - \\
    & CCVPE~\cite{xia2023convolutional} & 3.60 & 1.36 & - & - & 4.97 & 1.68 & - & - \\
    & GGCVT~\cite{shi2023boosting} & 4.12 &  1.34 & - & - & 5.16 & {1.40} & - & - \\
    & DenseFlow~\cite{song2024learning} & 3.03 & \textbf{0.97} & - & - & 5.01 & 2.42 & - & - \\
    & HC-Net~\cite{wang2024fine} & 2.65 & 1.17 & - & - & 3.35 & {1.59} & - & - \\
    & Ours & \textbf{1.95} & {1.08} & - & - & \textbf{2.41} & \textbf{1.37} & - & - \\
    \hline
    \multirow{5}{*}{Unknown} & 
    SliceMatch~\cite{lentsch2023slicematch} & 6.49 & 3.13 & 25.46 & 4.71 & 7.22 & 3.31 & 25.97 & 4.51 \\
    & CCVPE~\cite{xia2023convolutional} & \textbf{3.74} & \textbf{1.42} & 12.83 & 6.62 & \textbf{5.41} & \textbf{1.89} & 27.78 & 13.58 \\
    & DenseFlow~\cite{song2024learning} & 4.97 & 1.90 & \textbf{11.20} & 1.59 & 7.67 & 3.67 & \textbf{17.63} & 2.94 \\
    & Ours & 8.95 & 7.32 & 15.02 & 2.94 & 10.02 & 8.14 & 31.41 & 5.45 \\
    & Ours$^\dagger$ & 3.78 & 1.70 & {12.63} & \textbf{1.44} & 5.95 & 2.40 & 28.41 & \textbf{2.20} \\
    \bottomrule
    \end{tabular}
    \caption{VIGOR test results. Ours$^\dagger$ means using our method with a two-step inference to boost the performance. 
    \textbf{Best in bold.}}
    \label{tab:vigor}
\end{table*}

\textbf{VIGOR:}
We compare our method against previous state-of-the-art, including both global descriptor-based~\cite{xia2023convolutional, lentsch2023slicematch} and local feature-based~\cite{shi2023boosting, wang2024fine, song2024learning} methods.

As shown in Tab.~\ref{tab:vigor}, our method achieves a considerably lower mean localization error than others when the orientation is known. 
Compared to the previous state-of-the-art, HC-Net~\cite{wang2024fine}, our method reduces the mean localization error by 26\% and 28\% on the same-area and cross-area test sets, respectively.
Our median errors are on par with those of previous best-performing methods, DenseFlow~\cite{song2024learning} and GGCVT~\cite{shi2023boosting}, while having substantially lower mean errors.

Unknown orientation poses a greater challenge. 
GGCVT and HC-Net did not report results for this setting. 
The global descriptor-based method CCVPE~\cite{xia2023convolutional} outperforms all local feature-based methods, including ours.
CCVPE has a localization branch and an orientation branch.
Its localization branch is designed to be invariant to different orientations.
While our localization and orientation are computed jointly.
When orientation is unknown, our method must distinguish a tree along one viewing direction from all other trees in the scene, posing a challenging learning task. 
In contrast, when orientation is known, our model can focus on distinguishing objects along specific rays.

In practice, unknown orientation is less common, as we can obtain orientation estimates from a phone’s built-in compass or temporal information.
Nevertheless, we emphasize that, since our method accurately estimates orientation, we can use the model’s prediction to rotate the ground panorama to a roughly orientation-aligned setting. 
We then apply another model, trained with a small orientation noise range, e.g., $\pm20^\circ$, to estimate the pose again. 
This simple two-step inference 
largely reduces the gap to state-of-the-art when testing with unknown orientation.

\begin{table*}[h]
    \centering
    \begin{tabular}{p{0.25cm}p{2.5cm}p{0.8cm}p{1.1cm}p{1cm}p{1cm}p{1cm}p{1cm}p{0.8cm}p{1.1cm}p{0.88cm}p{0.97cm}}
    \toprule
    \multicolumn{2}{c}{\multirow{2}{*}{KITTI}} & \multicolumn{2}{c}{$\downarrow$ 
 Loc. (m)} & \multicolumn{2}{c}{$\uparrow$ Lateral ($\%$)} & \multicolumn{2}{c}{$\uparrow$ Long. ($\%$)} & \multicolumn{2}{c}{$\downarrow$ Orien. ($^\circ$)} & \multicolumn{2}{c}{$\uparrow$ 
 Orien. ($\%$)}\\
    \cline{3-12} 
    & & Mean & Median & R@1m & R@5m & R@1m & R@5m & Mean & Median & R@$1^\circ$ & R@$5^\circ$\\
    \hline
    \multirow{5}{*}{\rotatebox{90}{Same-area}} & GGCVT~\cite{shi2023boosting} & - &  - & 76.44 & 98.89 & 23.54 & 62.18 & - & - & \textbf{99.10} & \textbf{100.00}\\
    & CCVPE~\cite{xia2023convolutional} & 1.22 & 0.62 & 97.35 & 99.71 & 77.13 & 97.16 & 0.67 & 0.54 & 77.39 & 99.95 \\
    & HC-Net~\cite{wang2024fine} & {0.80} & 0.50 & 99.01 & 99.73 & \textbf{92.20} & \textbf{99.25} & \textbf{0.45} & {0.33} & 91.35 & 99.84 \\
    & DenseFlow~\cite{song2024learning} & 1.48 & \textbf{0.47} & 95.47 & 99.79 & 87.89 & 94.78 & 0.49 & \textbf{0.30} & 89.40 & 99.31 \\
    & Ours & \textbf{0.75} & 0.52 & \textbf{99.73} & \textbf{100.00} & 86.99 & 98.75 & 1.28 & 0.74 & 61.17 & 95.65 \\
    \hline
    \multirow{5}{*}{\rotatebox{90}{Cross-area}} & GGCVT~\cite{shi2023boosting} & - & - & 57.72 & 91.16 & 14.15 & 45.00 & - & - & \textbf{98.98} & \textbf{100.00}\\
    & CCVPE~\cite{xia2023convolutional} & 9.16 & \textbf{3.33} & 44.06 & 90.23 & 23.08 & 64.31 & \textbf{1.55} & \textbf{0.84} & 57.72 & 96.19 \\
    & HC-Net~\cite{wang2024fine} & 8.47 & 4.57 & 75.00 & 97.76 & \textbf{58.93} & \textbf{76.46} & 3.22 & 1.63 & 33.58 & 83.78\\
    & DenseFlow~\cite{song2024learning} & 7.97 & 3.52 & 54.19 & 91.74  & 23.10 & 61.75 & 2.17 & 1.21 & 43.44 & 89.31 \\
    & Ours & \textbf{7.45} & 4.03 & \textbf{89.46} & \textbf{99.80} & 12.42 & 55.73 & 3.33 & 1.88 & 30.34 & 81.17\\
    \bottomrule
    \end{tabular}
    \caption{KITTI test results. We did not include the almost pixel-perfect (KITTI aerial image's GSD: $1~\text{pixel} = 0.2~\text{m}$) localization result from~\cite{wang2024view}, as we cannot reproduce it because of the unavailability of the code.
    \textbf{Best in bold.}}
    \label{tab:kitti}
\end{table*}

\textbf{KITTI:}
Our method outperforms the previous state-of-the-art w.r.t. mean localization errors, while having slightly higher median errors, see Tab.~\ref{tab:kitti}.
Compared to HC-Net~\cite{wang2024fine}, our method achieves comparable performance when testing on images from the same area but shows much lower mean and median errors when testing on images from different areas, demonstrating better generalization capability.

Since KITTI images have a limited field of view, localization in the longitudinal direction is more challenging than that in the lateral direction. 
Consequently, all methods have high recall in the lateral direction, and ours achieves the highest.
For orientation prediction, our method performs slightly worse than the baselines. 
Most KITTI images view along the direction of the road,
which creates a bias that other methods can easily learn.
As the full aerial feature directly contributes to their output, simply predicting the road orientation leads to very low orientation errors.
In contrast, our method must infer the pose from sparse matching results, making it more challenging to exploit this bias.

\begin{figure*}[ht]
    \captionsetup[subfigure]{labelformat=empty}
    \tikzset{inner sep=0pt}
    \setkeys{Gin}{width=0.49\textwidth}
    \centering
    \subfloat[\label{fig:feature_matching_a}]{%
    \tikz{\node (a) {\includegraphics{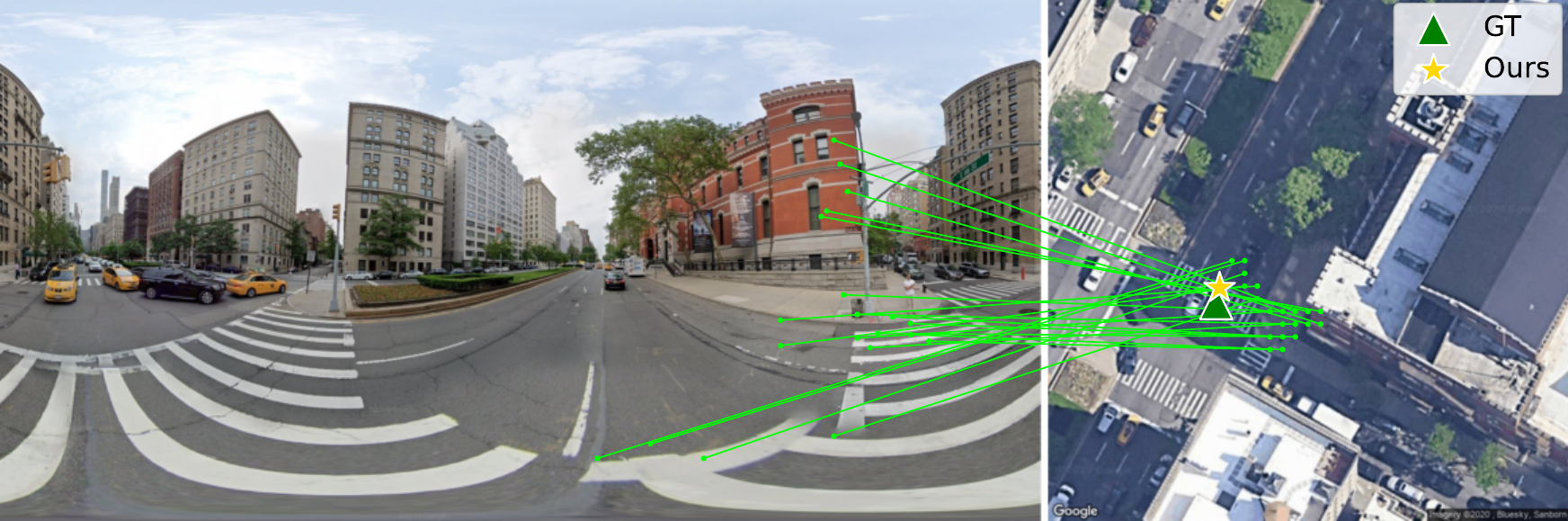}};
        \node[below right=2mm] at (a.north west) {(a)}; 
          }}
    \hfil
    \subfloat[\label{fig:feature_matching_b}]{%
    \tikz{\node (a) {\includegraphics{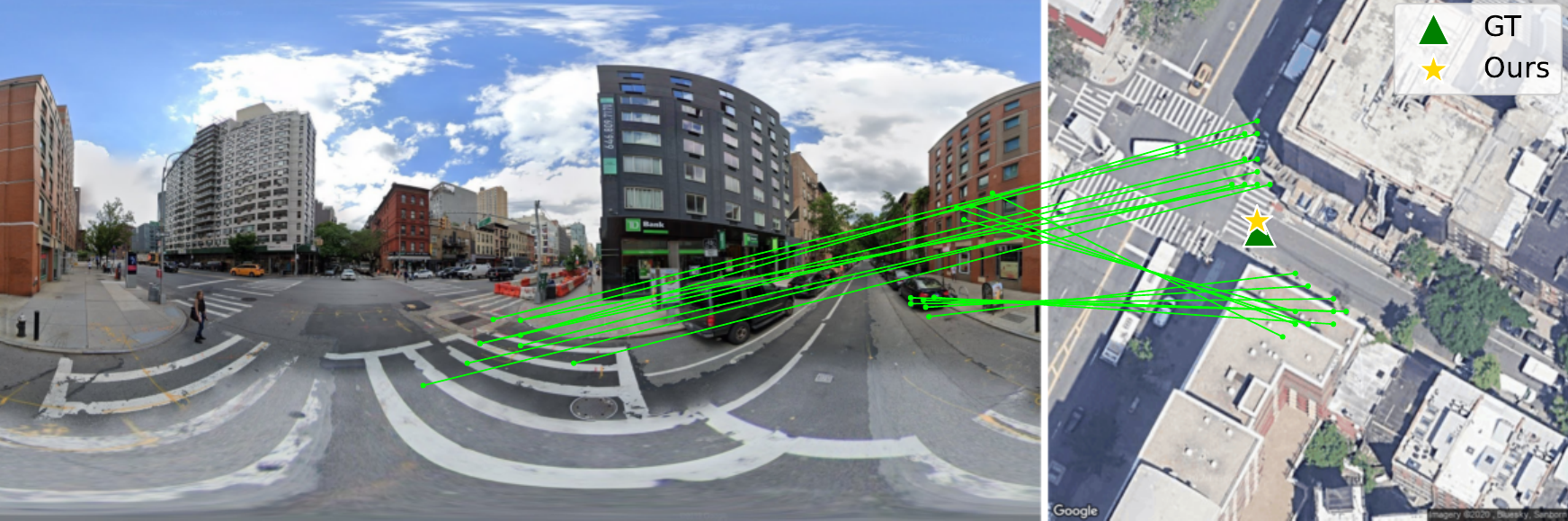}};
        \node[below right=2mm] at (a.north west) {(b)}; 
      }}
    \\
    \subfloat[\label{fig:feature_matching_c}]{%
    \tikz{\node (a) {\includegraphics{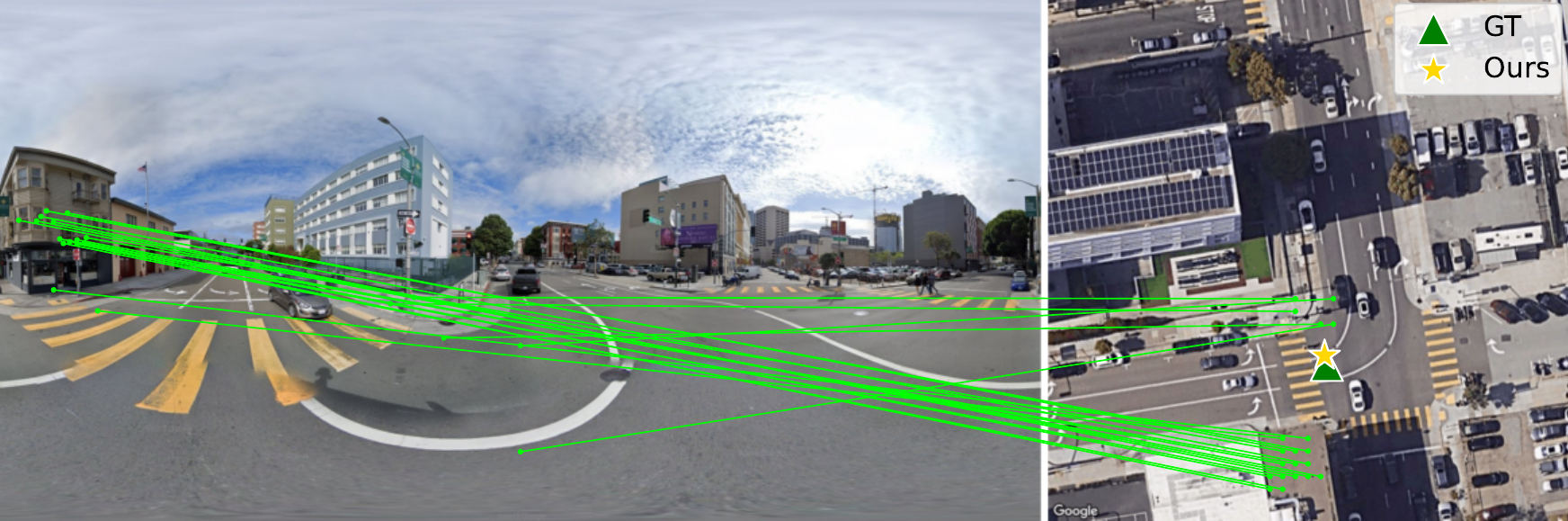}};
        \node[below right=2mm] at (a.north west) {(c)}; 
          }}
    \hfil
    \subfloat[\label{fig:feature_matching_d}]{%
    \tikz{\node (a) {\includegraphics{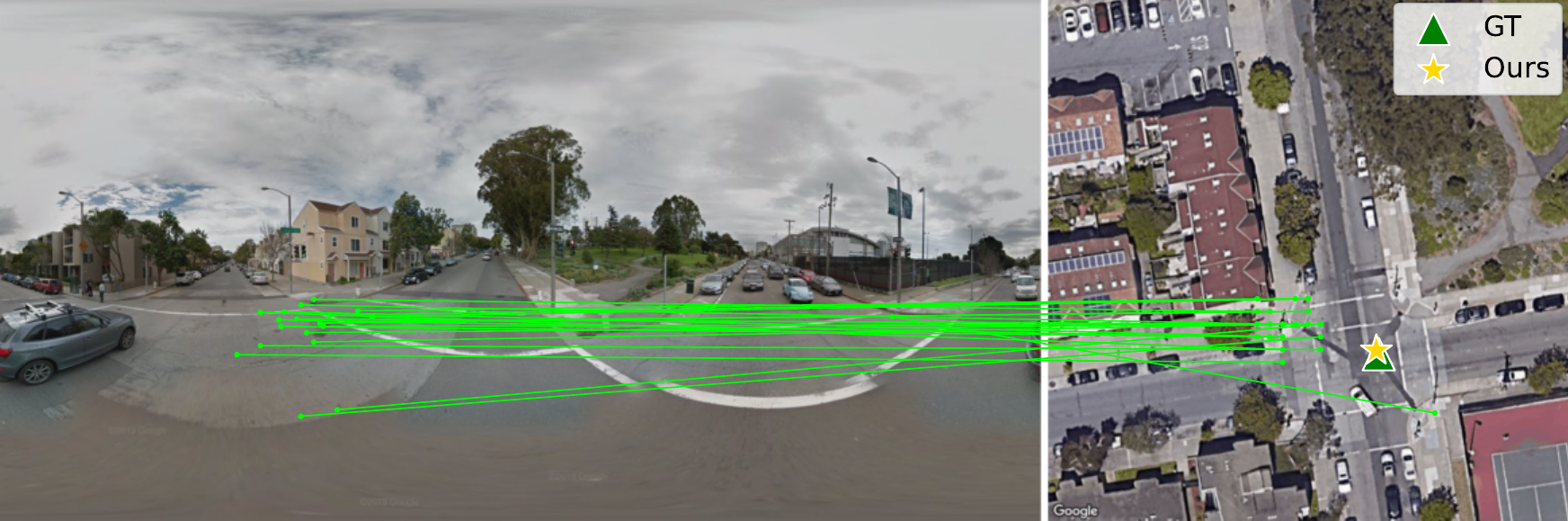}};
        \node[below right=2mm] at (a.north west) {(d)}; 
      }}
    \\\subfloat[\label{fig:feature_matching_e}]{%
    \tikz{\node (a) {\includegraphics{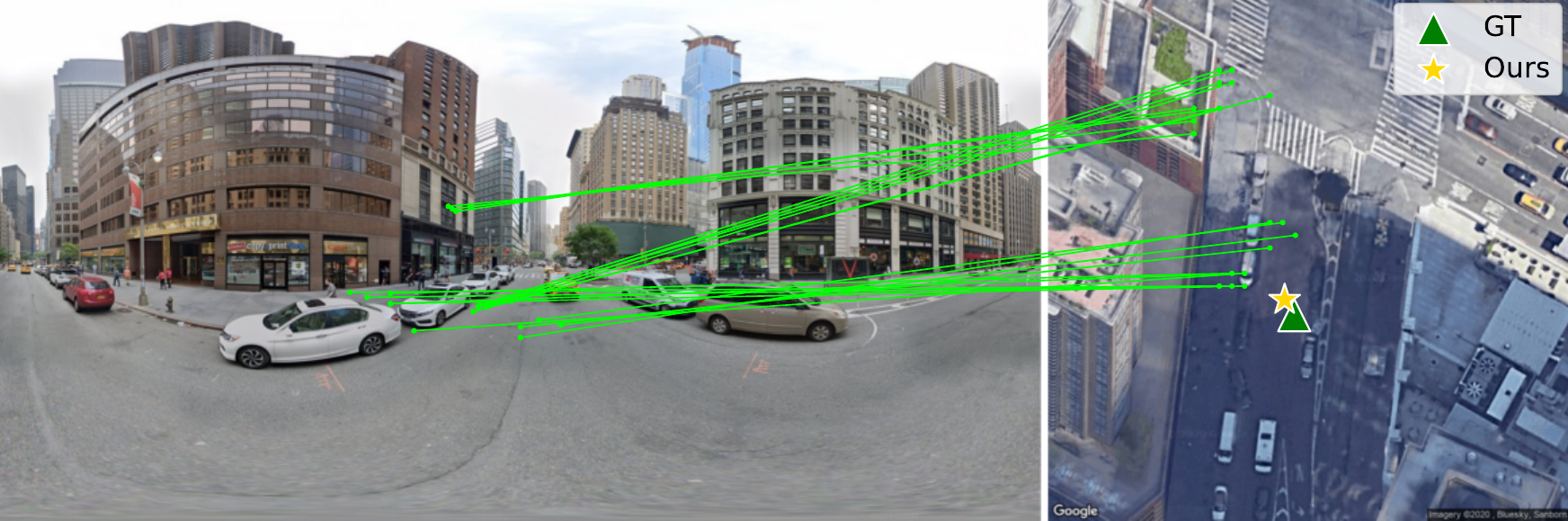}};
        \node[below right=2mm] at (a.north west) {(e)}; 
          }}
    \hfil
    \subfloat[\label{fig:feature_matching_f}]{%
    \tikz{\node (a) {\includegraphics{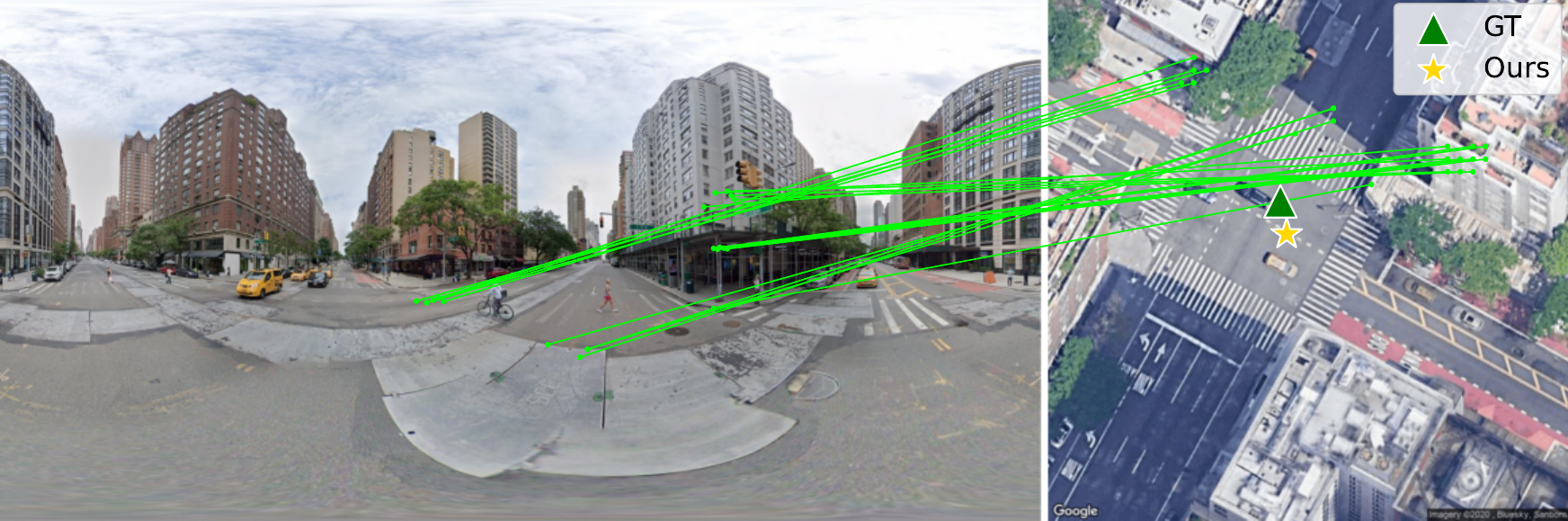}};
        \node[below right=2mm] at (a.north west) {(f)}; 
      }}
    \caption{Fine-grained feature matching results on VIGOR with known orientation. We show the 20 matches with the highest similarity scores. We find the 3D points using the selected height in the last pooling to BEV step and then project those points to the ground image.}
    \label{fig:feature_matching}
\end{figure*}

\begin{figure*}[ht]
    \centering
    \begin{minipage}{\linewidth}
    \centering
    \includegraphics[width=0.31\linewidth]{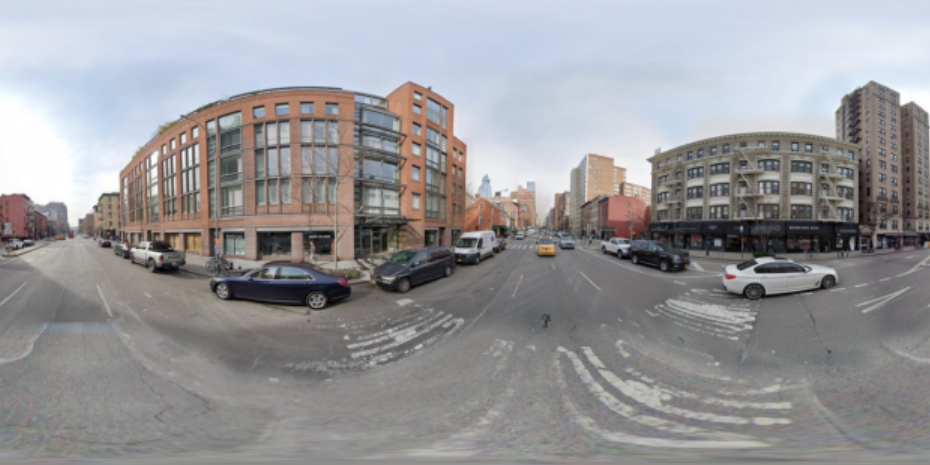}
    \includegraphics[width=0.18\linewidth]{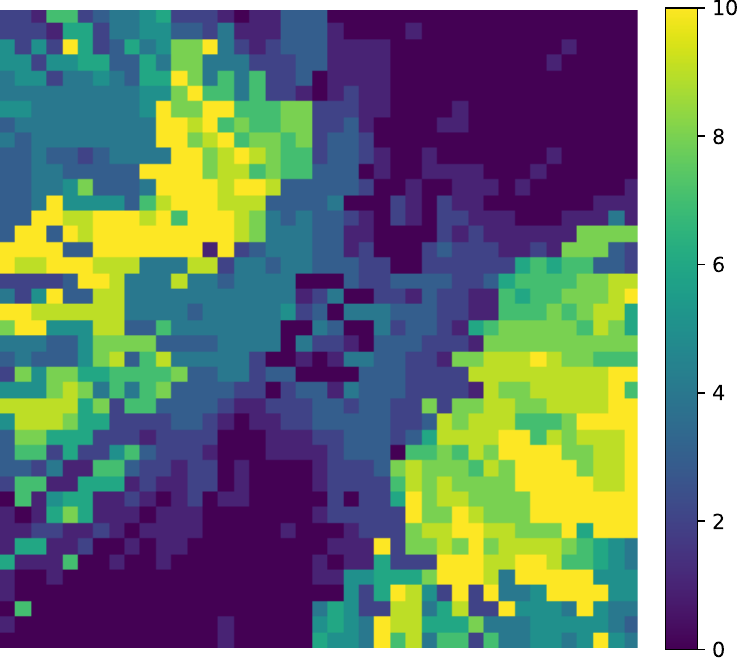} 
    \includegraphics[width=0.31\linewidth]{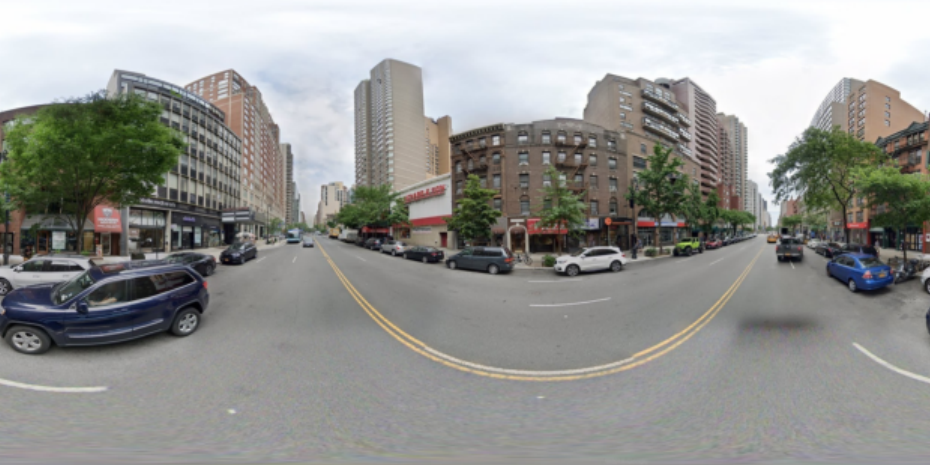}
    \includegraphics[width=0.18\linewidth]{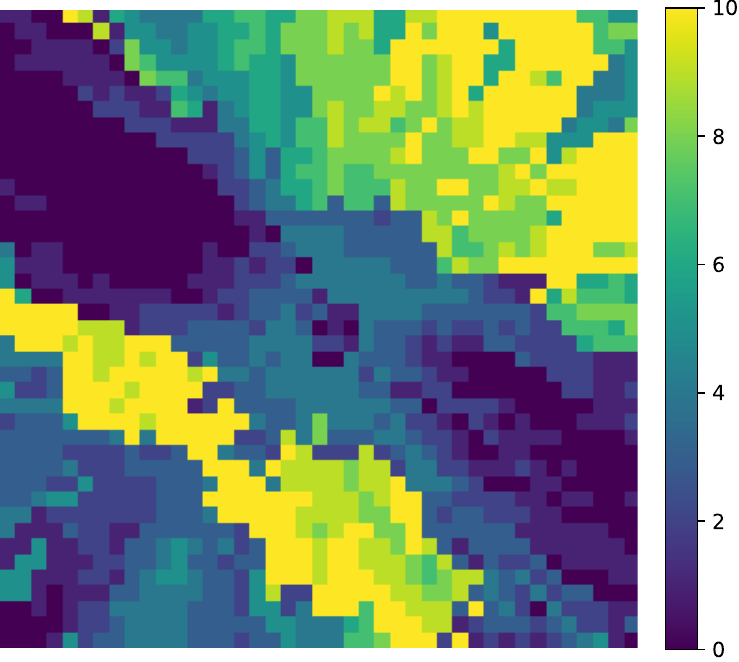} \\ 
    \end{minipage}
    \caption{Visualization of selected height indexes when pooling the ground 3D points to BEV. Each cell in the BEV map corresponds to a pillar of 3D points, and the color denotes the selected height (larger values mean higher points). The ground camera is located at the center of the BEV height map. The center vertical line in the panorama corresponds to a line originating from the center of the BEV to the top.}
    \label{fig:height_selection}
\end{figure*}

\subsection{Qualitative results}

\textbf{Fine-grained feature matching:}
A major advantage of our method is its interpretability. 
Global descriptor-based methods~\cite{xia2023convolutional, lentsch2023slicematch} condense the entire ground image into a single descriptor, making it unclear which ground local feature matches with which aerial local feature.
Previous local feature-based methods~\cite{song2024learning, wang2024fine} match features in BEV space, but they do not explicitly track which ground feature along the height contributes to the BEV representation. 
More importantly, they use a learned module to output the camera pose based on the matching scores. 
We observed with HC-Net~\cite{wang2024fine} that if we swap the order of ground and aerial features, its pose estimator does not produce an inverse pose of the unswapped version. 
This makes it unclear how feature matching influences the localization result.

In contrast, our estimated pose is a direct result of the sampled matching scores. 
Meanwhile, our proposed feature selection along height allows us to track which ground features in the image contribute to the matching in BEV space, making localization more interpretable.

We visualize the top 20 matched ground-aerial correspondences in Fig.~\ref{fig:feature_matching}.
Overall, the matched content is semantically consistent.
In (a) and (b), the points around zebra crossings in the ground view are matched to the zebra crossings in the aerial view.
In (c) and (d), the points near the curbs and road markings in the ground view are matched to those curbs in the aerial view.
Given that ground and aerial images capture the scene from perpendicular perspectives, our model matches building facades in the ground image to roofs in the aerial view, see Fig.~\ref{fig:feature_matching} (b), (c), (e) and (f), similar to how humans would do when localizing themselves. 
We emphasize again that this matching is learned in a weakly supervised manner, as we have only the ground truth camera pose and no explicit local feature correspondences.

\textbf{Visualization of selected height:}
Our fine-grained feature matching is enabled by the proposed feature selection along height. 
Next, we visualize the selected height when pooling the ground 3D points onto the BEV plane in Fig.~\ref{fig:height_selection}.

The first panorama shows a road in a roughly northeast-southwest direction, with buildings on the side.
The height map shows that the model focuses on lower-height features along the road while paying attention to higher points at the sides of the road.
Similarly, in the second example, the road runs from northwest to southeast, and our model again focuses on on-ground features along the road and shifts focus to higher points where there are buildings.
By supervising only the camera pose, our model learns to select appropriate features along the height based on the image content.

\subsection{Ablation study}
Next, an extensive ablation study is conducted on the VIGOR same-area validation set with known orientation to investigate our key hyperparameters:
the number of ground and aerial points $N$, the weight $\beta$ on our matching loss $\mathcal{L}_{M}$, the number of sampled correspondences $N_S$ for Procrustes alignment, and the effect of replacing our proposed selection along height with summing or taking the max feature.

As expected, our method benefits from a larger number of ground and aerial points $N$. 
We observe an increase in localization performance when increasing $N$ from $21 \times 21$ to $41 \times 41$, and the performance saturates after $41 \times 41$, see Tab.~\ref{tab:ablation}.
Thus, we use $N = 41 \times 41$.
When excluding our matching loss $\mathcal{L}_{M}$ by setting $\beta$ to 0, we observe a performance drop compared to our default setting.
Besides, the model converges more slowly without direct supervision on descriptor matching.
When $\beta$ is too large, i.e., $\beta=10$,  the supervision signal on the predicted pose becomes less dominant, resulting in lower localization accuracy.

In~\cite{fervers2023uncertainty,shi2023boosting}, the pose is inferred based on the dense matching result of ground and aerial features. 
Similarly, we tested using all point correspondences, incorporating their matching scores as weights for Procrustes alignment. 
However, this leads to higher errors, see ``No sampling'' in Tab.~\ref{tab:ablation}. 
Since matchable information between ground and aerial images is sparse, not all correspondences should contribute to pose estimation. 
Although unmatched correspondences may receive lower weights, a large number of such correspondences can still negatively impact accuracy.
We find our localization accuracy is not significantly influenced by $N_S$ when testing $N_S = 64, 128, 256, 512, 1024$, except that the model with $N_S = 64$ performs slightly worse. 
However, a larger $N_S$ results in a smoother training loss curve.

Instead of using our proposed feature selection when pooling the 3D points to BEV, we also tried simply summing~\cite{li2022bevformer} or taking the maximum~\cite{sarlin2024snap} feature over the 3D points along the height. 
As shown in the last two rows of Tab.~\ref{tab:ablation}, both formulations yield worse localization accuracy. 
Moreover, since the sum and max operations lose track of which features contribute to the BEV, they reduce the interpretability of the model.

\begin{table}[ht]
    \centering
    \begin{tabular}{p{1.3cm}p{2.3cm}p{1.3cm}p{1.3cm}}
    \toprule
    \multicolumn{2}{c}{Hyperparameters} & \multicolumn{2}{c}{$\downarrow$ Localization (m)} \\
    $N$ & Settings & Mean & Median  \\
    \hline
    $21 \times 21$ & default & 2.50 & 1.48\\
    $31 \times 31$ & default & 2.26 & 1.25 \\
    $41 \times 41$ & default & \textbf{2.17} & 1.18 \\
    $51 \times 51$ & default& 2.19 & \textbf{1.16} \\
    $41 \times 41$ & $\beta=0$  & 2.63 & 1.45 \\
    $41 \times 41$ & $\beta=10$ & 2.54 & 1.38 \\
    $41 \times 41$ & No sampling  & 2.55 & 1.45 \\
    $41 \times 41$ & sum over height & 2.34 & 1.27 \\
    $41 \times 41$ & max over height & 2.24 & 1.25 \\
    \bottomrule
    \end{tabular}
    \caption{Ablation study. Default: $\beta=1$, sampling with $N_S=1024$, and learning feature selection over height. \textbf{Best in bold.}}
    \label{tab:ablation}
\end{table}

\section{Conclusion}

We propose a novel fine-grained cross-view localization method that matches fine-grained features across views.
Our model learns semantically consistent matches in a weakly supervised manner, requiring only the camera pose for training.
Our method estimates the ground camera's pose by aligning a generated ground point plane with a sampled aerial point plane using Procrustes alignment.
This process involves mapping ground image features to a 3D point cloud and then pooling the 3D points into a BEV plane.
Instead of the standard approach of summing or taking the maximum feature along the height dimension when pooling to BEV, we proposed to learn the feature selection. 
This approach allows us to trace which ground image features contribute to the BEV representation and also improves localization accuracy.
Experimental results show that our method surpasses previous state-of-the-art methods by a large margin, reducing the mean localization error by 28\% on the challenging VIGOR cross-area test set.
When orientation is unknown, we propose a two-step inference: first, using one model to estimate the orientation and bring the input to a roughly orientation-aligned setting; then, using another model trained with small orientation noise for accurate pose estimation. 
This approach achieves performance comparable to previous state-of-the-art methods.

\newpage
{
    \small
    \bibliographystyle{ieeenat_fullname}
    \bibliography{main}
}

\clearpage
\setcounter{page}{1}
\setcounter{table}{0} 
\setcounter{figure}{0}
\setcounter{equation}{0}
\maketitlesupplementary

\addcontentsline{toc}{section}{Appendices}
\appendix

In this supplementary material, we provide the following information to support the main paper:
\begin{enumerate}[label=\Alph*]
    \item Projection geometry.
    \item Procrustes alignment.
    \item Additional details about RANSAC.
    \item Additional comparison of loss functions.
    \item Assumption on orthographic images.
    \item Directly matching DINO feature.
    \item Extra Qualitative Results.
\end{enumerate}

\section{Projection geometry}
We provide details on how to find the projected pixel coordinates of the 3D points in the ground image feature map $f(G)$ during the mapping to 3D step in Sec.~3.2 of the main paper.

\begin{figure}[h]
    \centering
    \includegraphics[width=1\linewidth]{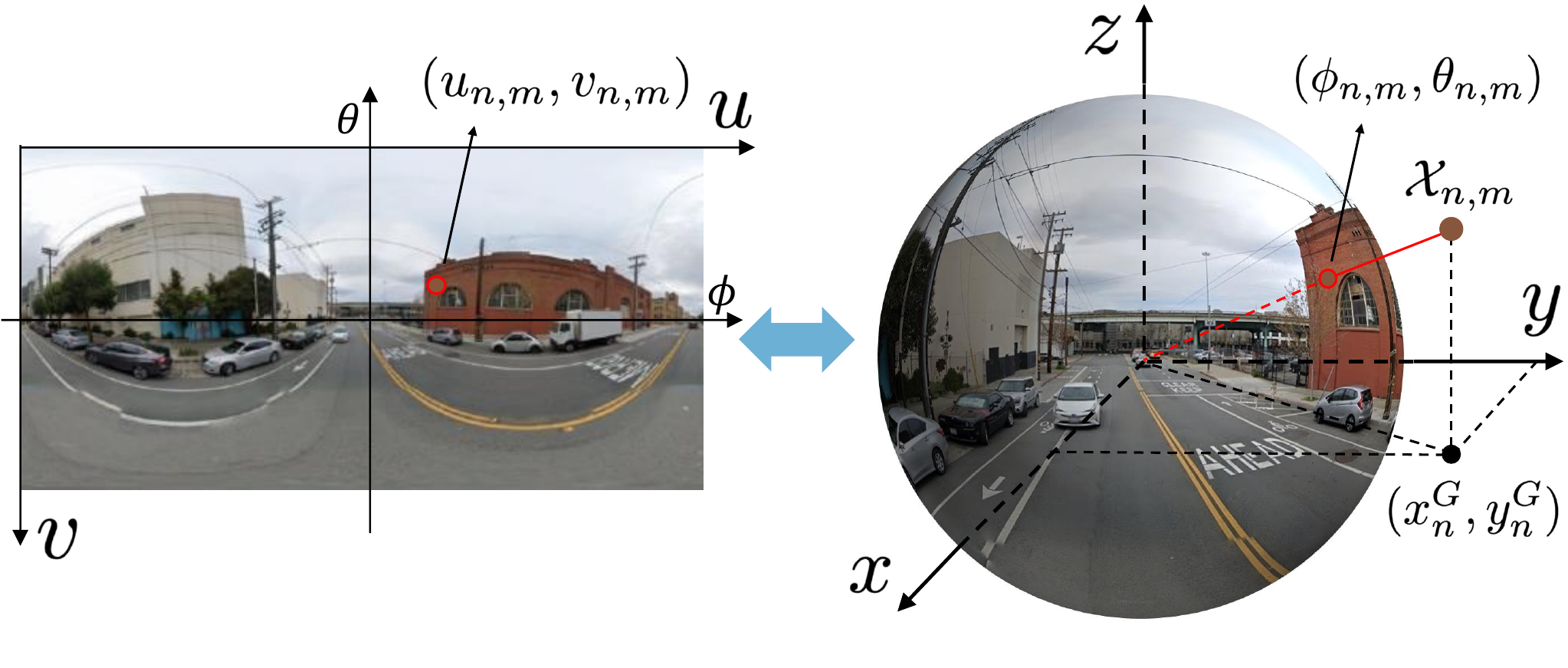}
    \caption{Projecting a 3D point to a panoramic image. We use the original image for visualization purposes. In practice, we find the projected pixel coordinates in the extracted feature map.}
    \label{fig:projection}
\end{figure}

We begin with the case where ground images are panoramic.
As shown in Fig.~\ref{fig:projection}, a panorama represents the surface of a sphere.
Each pixel in a panorama is associated with a spherical coordinate $(\phi, \theta)$, where $\phi \in [-\pi, \pi]$ and $\theta \in [-\frac{\pi}{2}, \frac{\pi}{2}]$.
Given the coordinate $(x^G_{n}, y^G_{n}, z^G_{n,m})$ of the 3D point $\mathcal{X}_{n,m}$, where $n$ is the index of the ground 2D point in the created point set $\xi^G$ and $m$ is the index of the lifted 3D point, we compute its projected spherical coordinates $(\phi_{n,m}, \theta_{n,m})$ using,

\begin{align}
    \phi_{n,m} = \begin{cases}
    \text{arccos}(\frac{x^G_n}{\sqrt{(x^G_n)^2 + (y^G_n)^2}}), & \text{if } y^G_n \geq 0 \\
    -\text{arccos}(\frac{x^G_n}{\sqrt{(x^G_n)^2 + (y^G_n)^2}}),              & \text{otherwise,} 
    \end{cases} \\
    \theta_{n,m} = \text{arcsin}(\frac{z^G_{n,m}}{\sqrt{(x^G_n)^2 + (y^G_n)^2 + (z^G_{n,m})^2}}).
\end{align}

Once we have $(\phi_{n,m}, \theta_{n,m})$, we can simply find its corresponding pixel coordinates $(u_{n,m}, v_{n,m})$,
\begin{align}
    u_{n,m} = \frac{\phi_{n,m} + \pi}{2 \pi} W, \\
    v_{n,m} = \frac{\frac{\pi}{2}-\theta_{n,m}}{\pi} H,
\end{align}
where $W$ and $H$ are the width and height of the extracted ground feature map $f(G)$.

\begin{figure}[h]
    \centering
    \includegraphics[width=1\linewidth]{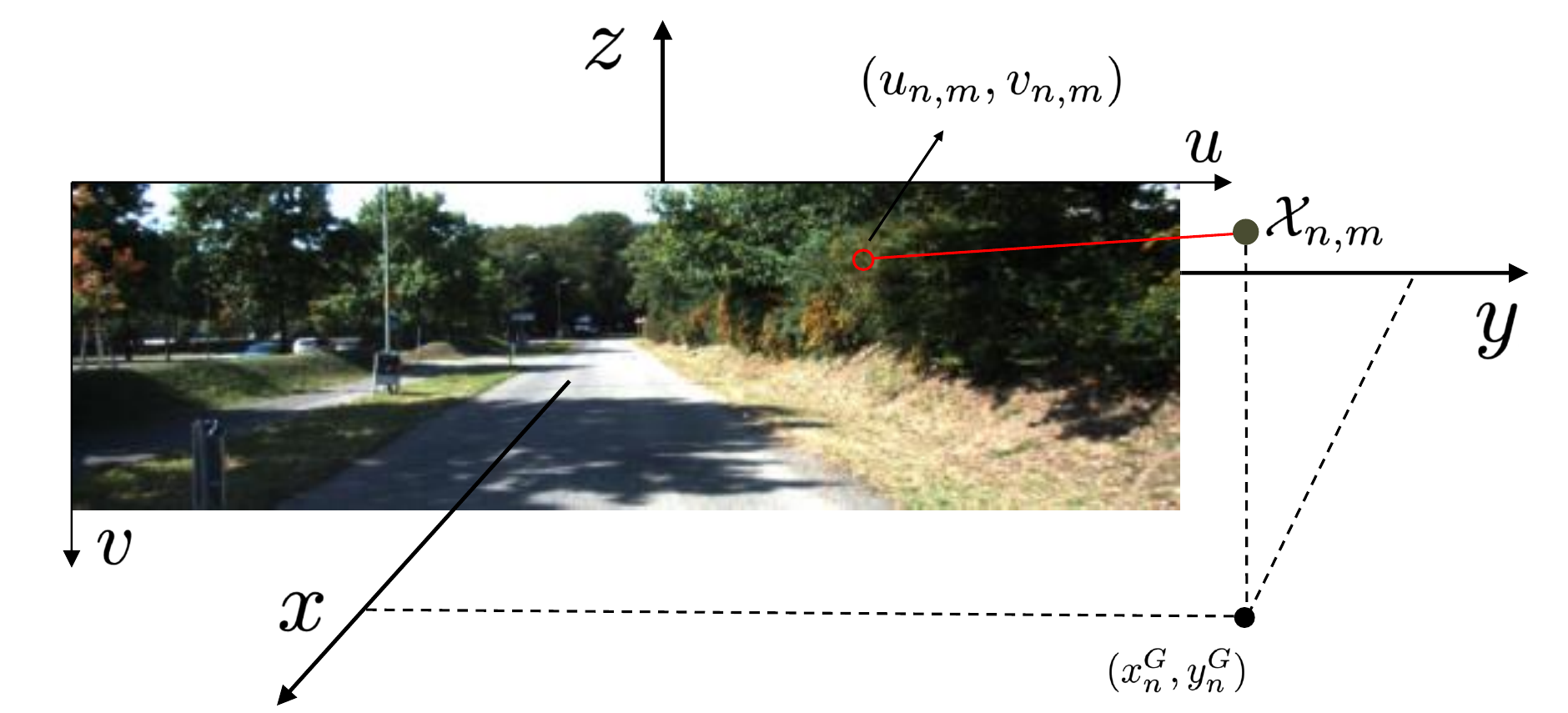}
    \caption{Perspective projection of a 3D point. We use the original image for visualization purposes. In practice, we find the projected pixel coordinates in the extracted feature map.}
    \label{fig:perspective_projection}
\end{figure}

If the ground images follow a perspective projection, see Fig.~\ref{fig:perspective_projection}, the pixel coordinates $(u_{n,m}, v_{n,m})$ of each 3D point $\mathcal{X}_{n,m} = (x^G_{n}, y^G_{n}, z^G_{n,m})$ can be computed using the camera intrinsics, 

\begin{align}
    u_{n,m} = \frac{W}{2} + \frac{\text{focal length} \cdot y^G_{n} }{x^G_{n} \cdot \text{stride}} + u_0, \\
    v_{n,m} = \frac{H}{2} - \frac{\text{focal length} \cdot z^G_{n,m} }{x^G_{n} \cdot \text{stride}} + v_0,
\end{align}
where \text{stride} is the downscale factor between the input image $G$ and its extracted feature map $f(G)$ and $(u_0, v_0)$ denotes the amount of principal point offset in the feature map $f(G)$.

\section{Procrustes alignment}
As mentioned in Sec.~3.3 of the main paper, we use the Kabsch solver~\cite{kabsch1976solution}, also known as the orthogonal Procrustes algorithm, to compute the relative pose between ground and aerial 2D point sets $\xi^G$ and $\xi^A$.
Kabsch solver is common for point cloud registration.

The Kabsch solver computes the optimal rotation matrix that minimizes the root mean squared deviation between two paired sets of points.
The translation between the two point sets is determined by calculating the shift between the resulting two rotation-aligned point sets.
This process is inherently differentiable~\cite{avetisyan2019end,brachmann2021visual}, as the rotation is obtained through the Singular Value Decomposition (SVD) of the cross-correlation matrix between the two point sets.

As noted in Sec.~3.3, given the $N$ ground and aerial 2D points $\xi^G$ and $\xi^A$ and their pair-wise matching probability $D$ (of length of $N^2$), we sample $N_S$ correspondences,
\begin{align}
    \{D_{n_S} \} \sim D, \quad n_S=1, ..., N_S,
\end{align}
where $n_S$ is the sampled index in $D$.
We denote the ground and aerial points of those $N_S$ correspondences as $\xi^G_S$ and $\xi^A_S$.
Then, the Kabsch solver computes the rotation (in our case, yaw orientation $o$) between $\xi^G_S$ and $\xi^A_S$ in three steps.

First, $\xi^G_S$ and $\xi^A_S$ are translated such that their centroid coincides with the origin of the coordinate system,
\begin{align}
    \xi^G_\text{trans} = \xi^G_S - \text{mean}(\xi^G_S), \quad
    \xi^A_\text{trans} = \xi^A_S - \text{mean}(\xi^A_S),
\end{align}
where $\xi^G_\text{trans}$ and $\xi^A_\text{trans}$ represent the translated ground and aerial points, respectively, and $\text{mean}(\cdot)$ computes the mean coordinate of the points in $\xi^G_S$ and $\xi^A_S$.
When computing the mean, the matching probabilities $\{D_{n_S} \}$ are used as weights~\cite{choy2020deep}, which means a point with a higher matching probability contributes more to the mean coordinate.

Then, we compute a cross-covariance matrix $H$ between $\xi^G_\text{trans}$ and $\xi^A_\text{trans}$ as $H = (\xi^A_\text{trans})^T \cdot \xi^G_\text{trans}$.

Finally, we compute the SVD of $H$,
\begin{align}
    U \cdot S \cdot V^T = H,
\end{align}
where $U$ and $V$ are orthogonal and $S$ is diagonal.
The $2 \times 2$ rotation matrix $rot$ of the orientation $o$ is then,
\begin{align}
    rot = U \cdot \begin{bmatrix}
    1 & 0  \\
    0 & \text{determinant}(U \cdot V^T) 
    \end{bmatrix} \cdot V^T.
\end{align}

Once the orientation is estimated, the translation $\boldsymbol{t}_{m}$ between $\xi^G_S$ and $\xi^A_S$ is computed as $\boldsymbol{t}_{m} = \text{mean}(\xi^G_S) - \text{mean}(\xi^A_S) \cdot rot^T$.

\section{Additional details about RANSAC}

RANSAC is not used during training.
When inference with RANSAC, we conduct $R=100$ iterations and use 2.5~m as the threshold in Eq.~5 in the main paper.
As shown in Tab.~\ref{tab:ransac}, on the VIGOR same-area test set, RANSAC reduces the mean localization error from 2.18~m to 1.95~m.
Without RANSAC, our method achieves 9.26 FPS on an H100 GPU.
As expected, RANSAC increases computation, and inference with RANSAC has 0.32 FPS.
If fast runtime is required, our method can be used without RANSAC, and it still surpasses the previous state-of-the-art in localization accuracy.
RANSAC is used to obtain the accurate pose estimation reported in the main paper (Tab.~1), whereas it is omitted in the ablation study (Tab.~3) due to runtime considerations.
On the KITTI dataset, we did not observe an overall performance improvement with RANSAC.
Therefore, we did not use RANSAC on KITTI.

\begin{table}[h]
    \centering
    \begin{tabular}{p{1.2cm}p{0.6cm}p{0.9cm}p{1.1cm}p{0.9cm}p{1.1cm}}
    \toprule
    \multirow{2}{*}{{VIGOR}} & \multirow{2}{*}{{FPS}} & \multicolumn{2}{c}{Same-area} & \multicolumn{2}{c}{Cross-area}  \\
    \cline{3-6} 
    & & Mean & Median & Mean & Median \\
    \hline
    w/o. R. & \textbf{9.26} & 2.18 & 1.18 & 2.74 & 1.52 \\
    w. R. & 0.32 & \textbf{1.95} & \textbf{1.08} & \textbf{2.41} & \textbf{1.37}  \\
    \bottomrule
    \end{tabular}
    \caption{Comparison of localization errors and runtime between our model with (``w. R.'') and without RANSAC (``w/o. R.'') when performing inference on the VIGOR test set with known orientation.
    \textbf{Best in bold.}}
    \label{tab:ransac}
\end{table}

\section{Additional comparison of loss functions}
We use a Virtual Correspondence Error loss, $\mathcal{L}_{VCE}$, adapted from~\cite{arnold2022map}, to supervise the camera pose (main paper Sec.~3.4). 
Here, we also compare it with other pose losses.

We tested a loss function that sums an L1 loss for localization and another L1 loss for orientation estimation, denoted as $\mathcal{L}_{A}$, and an alternative, 
$\mathcal{L}_{B}$, which replaces the L1 loss for localization with an L2 loss (i.e., minimizing squared distance).
As shown in Tab.~\ref{tab:loss_comparison}, both $\mathcal{L}_{A}$ and $\mathcal{L}_{B}$ performed worse than our $\mathcal{L}_{VCE}$.
$\mathcal{L}_{VCE}$ jointly considers localization and orientation, potentially leading to a better local optimum during training.

\vspace{-1mm}
\begin{table}[h]
    \centering
    \begin{tabular}{p{2cm}p{1.2cm}p{1.1cm}p{1.1cm}}
    \toprule
    Error (m) & $\mathcal{L}_{VCE}$ & $\mathcal{L}_{A}$ & $\mathcal{L}_{B}$ \\
    \hline
    Mean & \textbf{2.17} & 2.68 & 2.58 \\
    \hline
    Median & \textbf{1.18} & 1.38 & 1.37 \\
    \bottomrule
    \end{tabular}
    \caption{Loss comparison on VIGOR validation set with known orientation. \textbf{Best in bold.}}
    \label{tab:loss_comparison}
\end{table}
\vspace{-1mm}

\section{Assumption on orthographic images}
Our method assumes each pixel in the aerial image corresponds to a vertical ray in 3D (main paper Sec.~3.1).
However, existing cross-view localization datasets often violate this assumption.

We do not explicitly address non-orthographic aerial images. 
Since the model is supervised using the camera pose, or equivalently, the ground and aerial BEV point locations, it must learn to assign matchable features to corresponding ground and aerial points.
During training, building facades are sometimes visible in the aerial image, while at other times only the rooftops are seen. 
As a result, the model must learn to associate ground-view content (e.g., buildings) with both facades and roofs. 
This likely explains why, in the aerial view, points on buildings in the ground view are sometimes matched to roofs (main paper Fig.~3c) and other times to facades (main paper Fig.~3a).

On the other hand, since our point set is relatively sparse and the point descriptors are constructed from deep features, these descriptors likely summarize information from local neighborhoods. 
It is also possible that the descriptors for roof or facade points capture more global semantic information, e.g., identifying a region as part of a building, rather than strictly encoding the visual appearance of the projected location in the input images.

\section{Directly matching DINO feature}
\begin{figure*}[ht]
    \captionsetup[subfigure]{labelformat=empty}
    \tikzset{inner sep=0pt}
    \setkeys{Gin}{width=0.49\textwidth}
    \centering
    \subfloat[]{%
    \tikz{\node (a) {\includegraphics{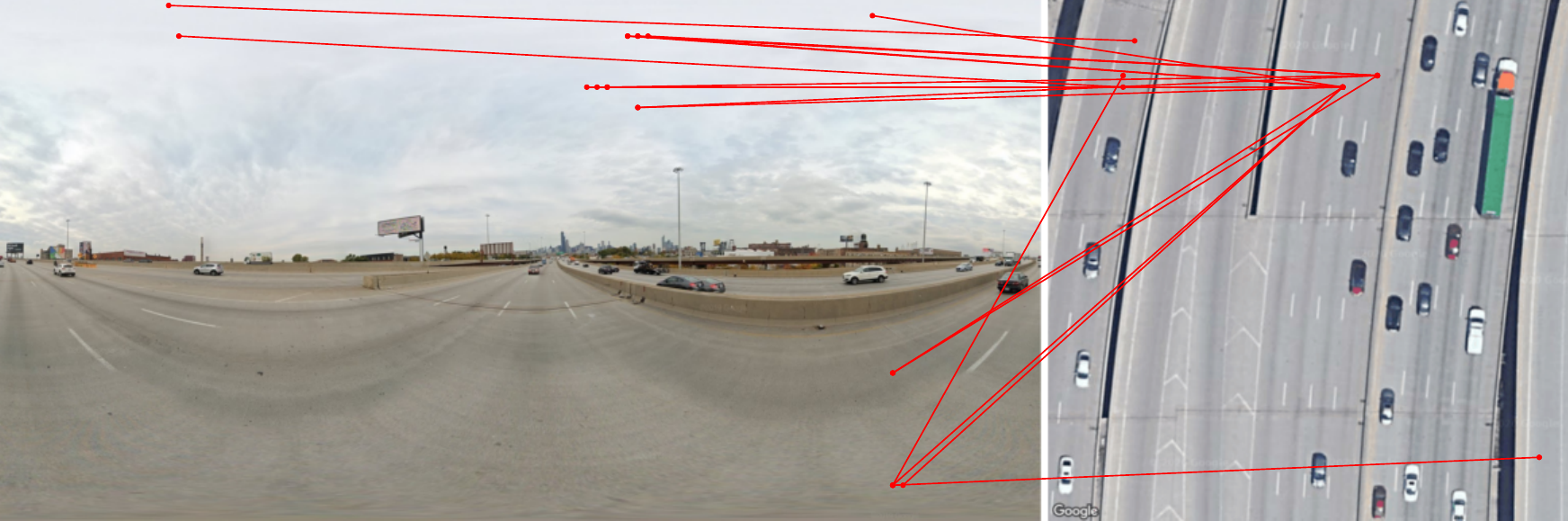}};
        \node[below right=2mm] at (a.north west) {(a)}; 
      }}
    \hfil
    \subfloat[]{%
    \tikz{\node (a) {\includegraphics{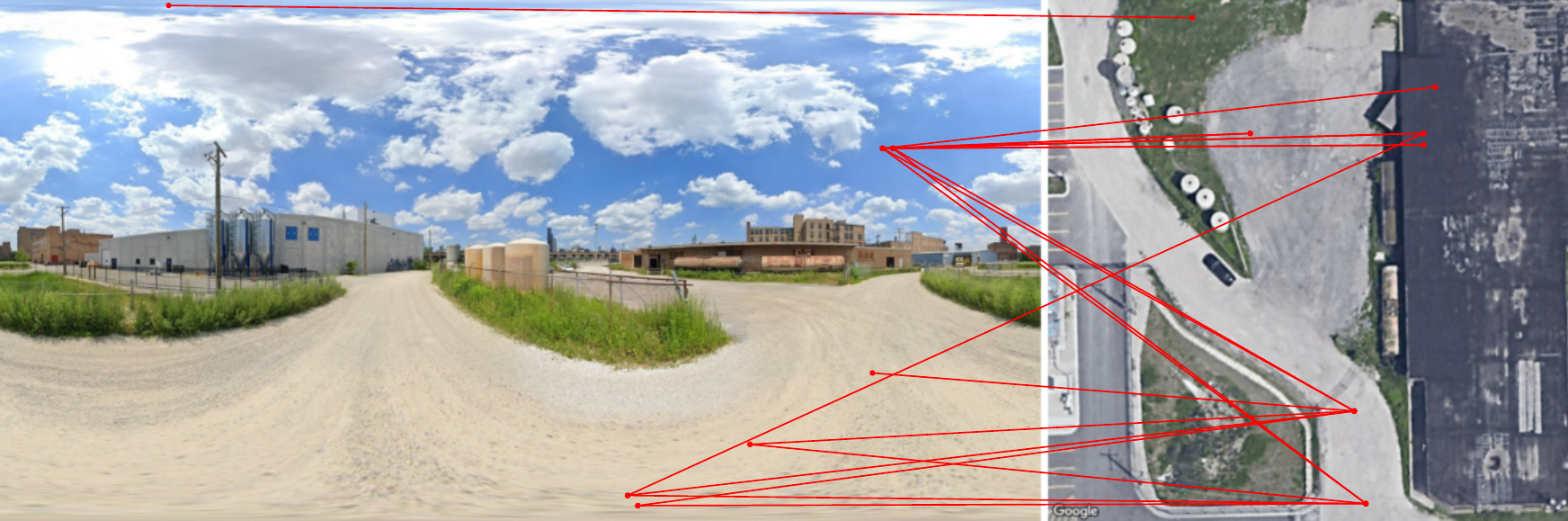}};
        \node[below right=2mm] at (a.north west) {(b)}; 
      }}
    \hfil
    \subfloat[]{%
    \tikz{\node (a) {\includegraphics{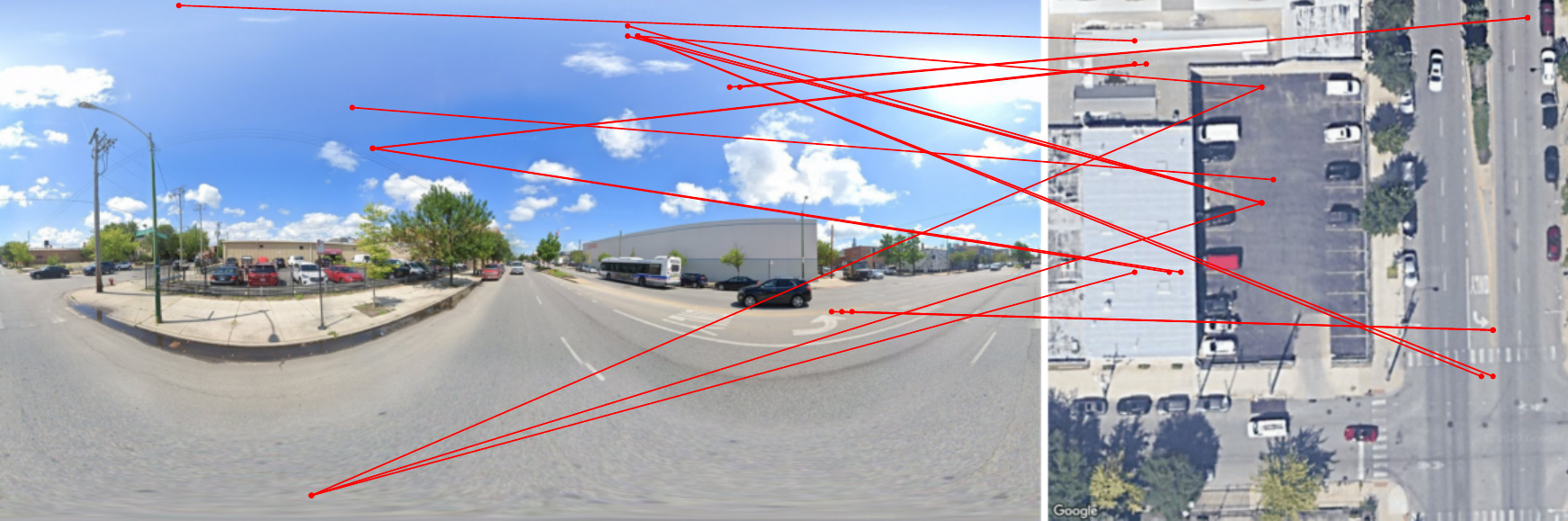}};
        \node[below right=2mm] at (a.north west) {(c)}; 
      }}
    \hfil
    \subfloat[]{%
    \tikz{\node (a) {\includegraphics{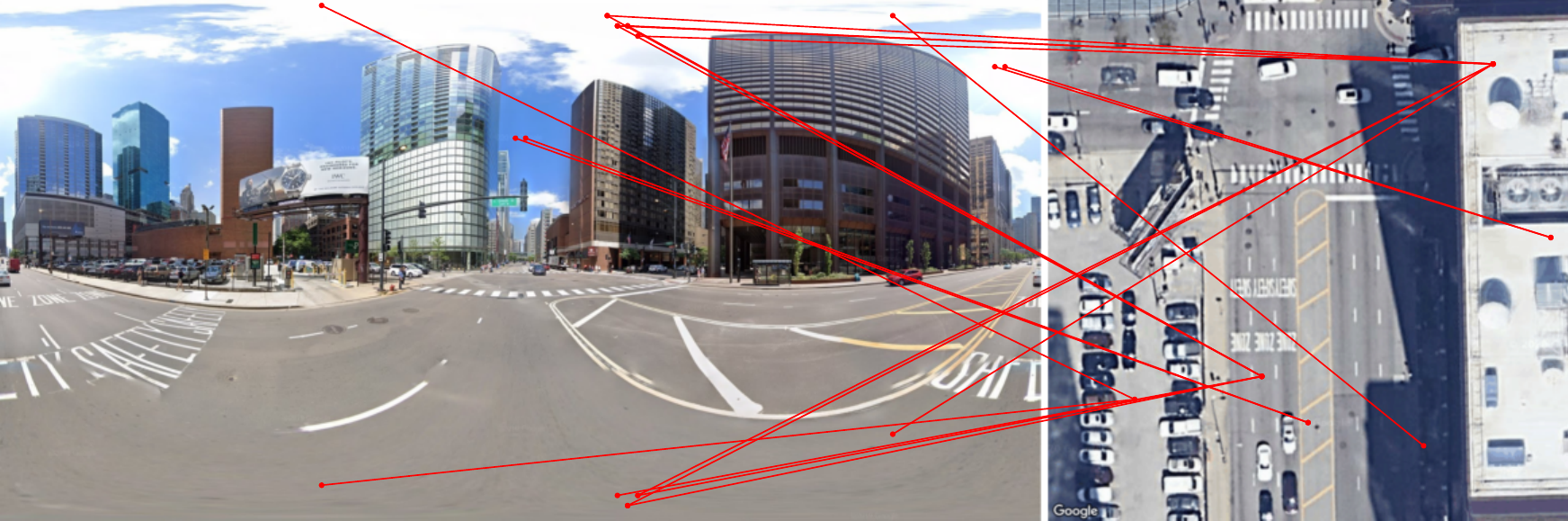}};
        \node[below right=2mm] at (a.north west) {(d)}; 
      }}
    \caption{Directly matching DINOv2 features~\cite{oquab2023dinov2} of the ground and aerial images. We show the 20 matches with the highest similarity scores.}
    \label{fig:dino}
\end{figure*}

As mentioned in the main paper, we use pre-trained DINOv2 features~\cite{oquab2023dinov2}. 
We tried directly matching DINO features between ground and aerial images.
However, this approach did not produce visually reasonable matches, see Fig.~\ref{fig:dino}.
Therefore, using our model on top of the DINO features is essential.

The DINOv2 paper~\cite{oquab2023dinov2} demonstrated impressive results by directly matching features extracted from images and sketches. 
However, ground-to-aerial cross-view image matching is significantly more challenging due to factors such as the lack of clear separation between foreground and background, the presence of dynamic and unmatchable objects, the repetition of structures like buildings and trees, and the substantial differences in perspective, scale, and scene coverage between the two views.

\section{Extra qualitative results}
\begin{figure*}[ht]
    \captionsetup[subfigure]{labelformat=empty}
    \tikzset{inner sep=0pt}
    \setkeys{Gin}{width=0.49\textwidth}
    \centering
    \subfloat[]{%
    \tikz{\node (a) {\includegraphics{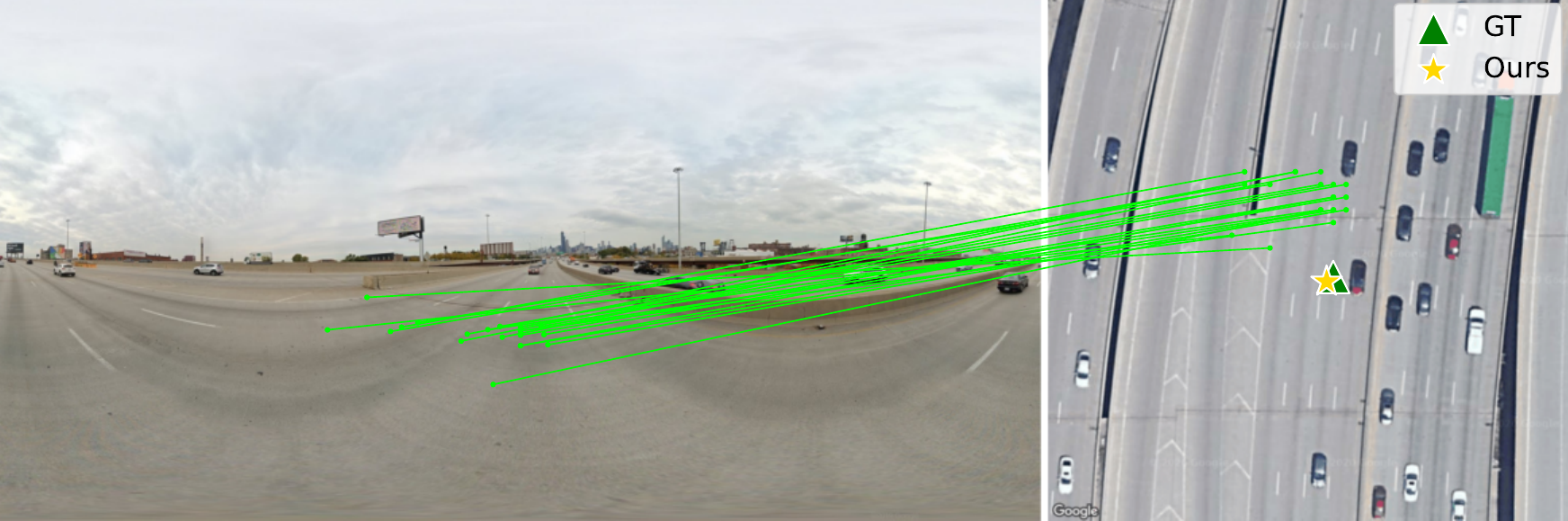}};
        \node[below right=2mm] at (a.north west) {(a)}; 
      }}
    \hfil
    \subfloat[]{%
    \tikz{\node (a) {\includegraphics{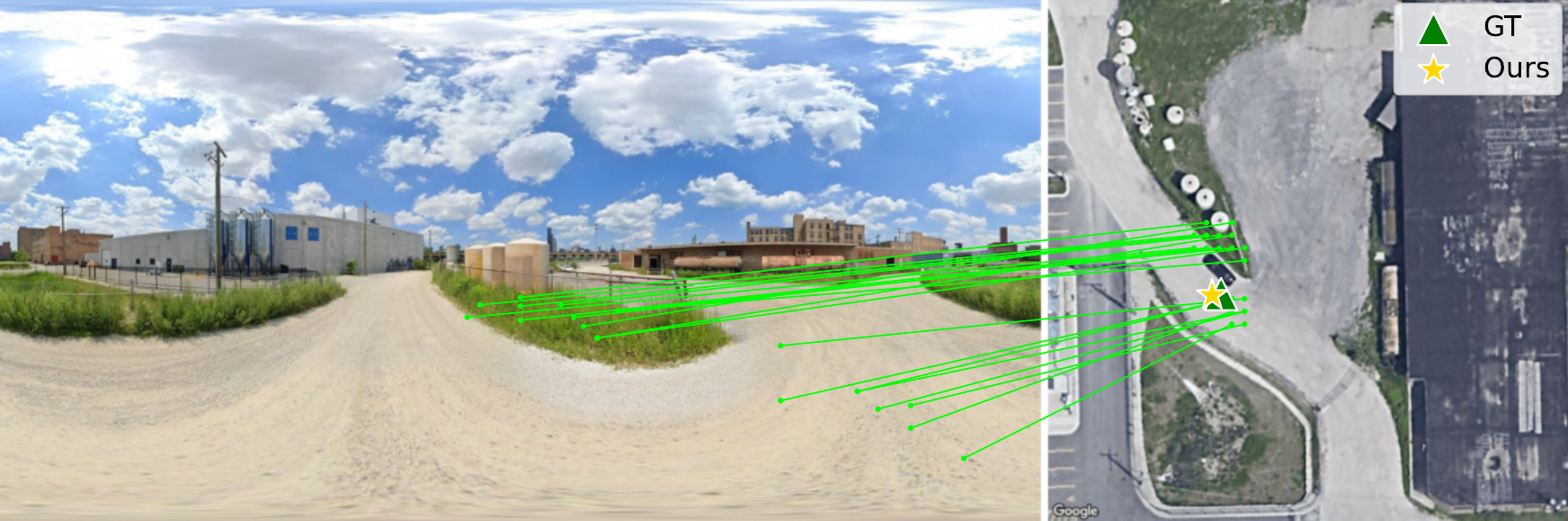}};
        \node[below right=2mm] at (a.north west) {(b)}; 
      }}
    \hfil
    \subfloat[]{%
    \tikz{\node (a) {\includegraphics{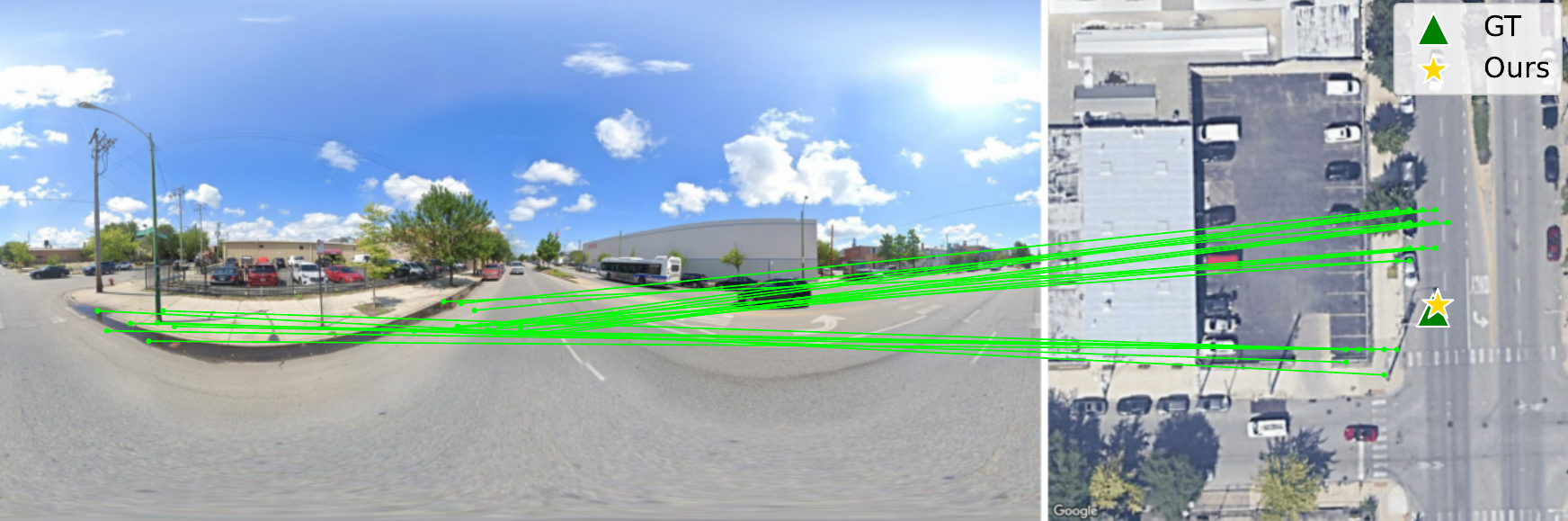}};
        \node[below right=2mm] at (a.north west) {(c)}; 
      }}
    \hfil
    \subfloat[]{%
    \tikz{\node (a) {\includegraphics{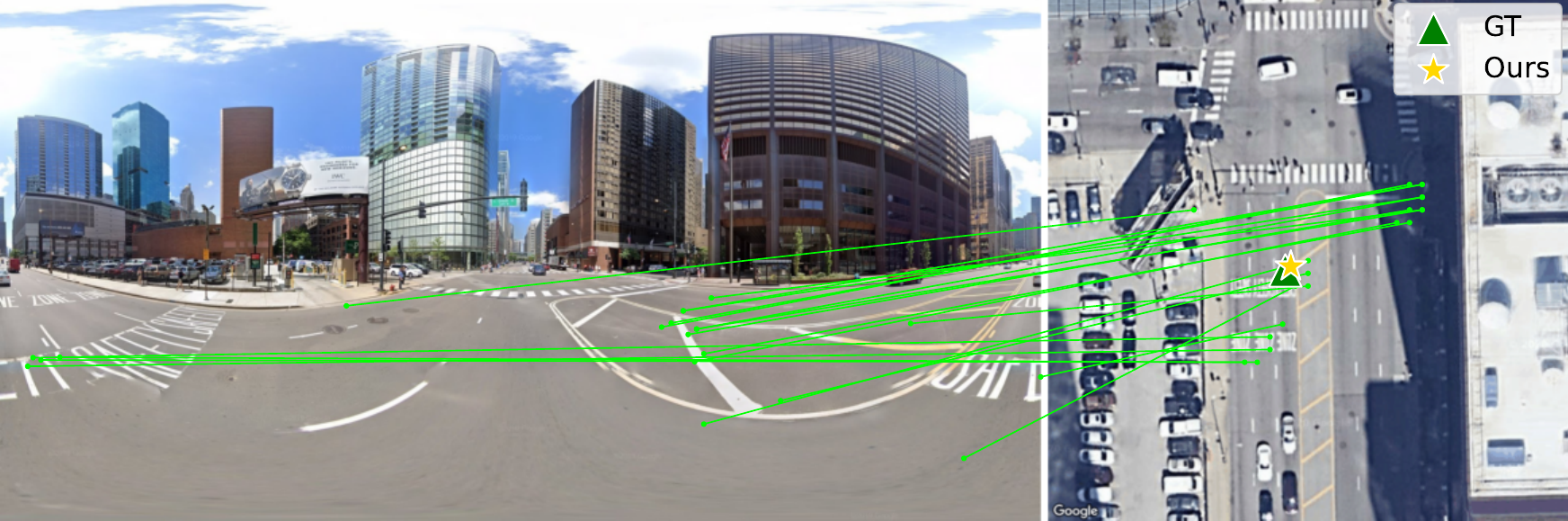}};
        \node[below right=2mm] at (a.north west) {(d)}; 
      }}
    \hfil
    \subfloat[]{%
    \tikz{\node (a) {\includegraphics{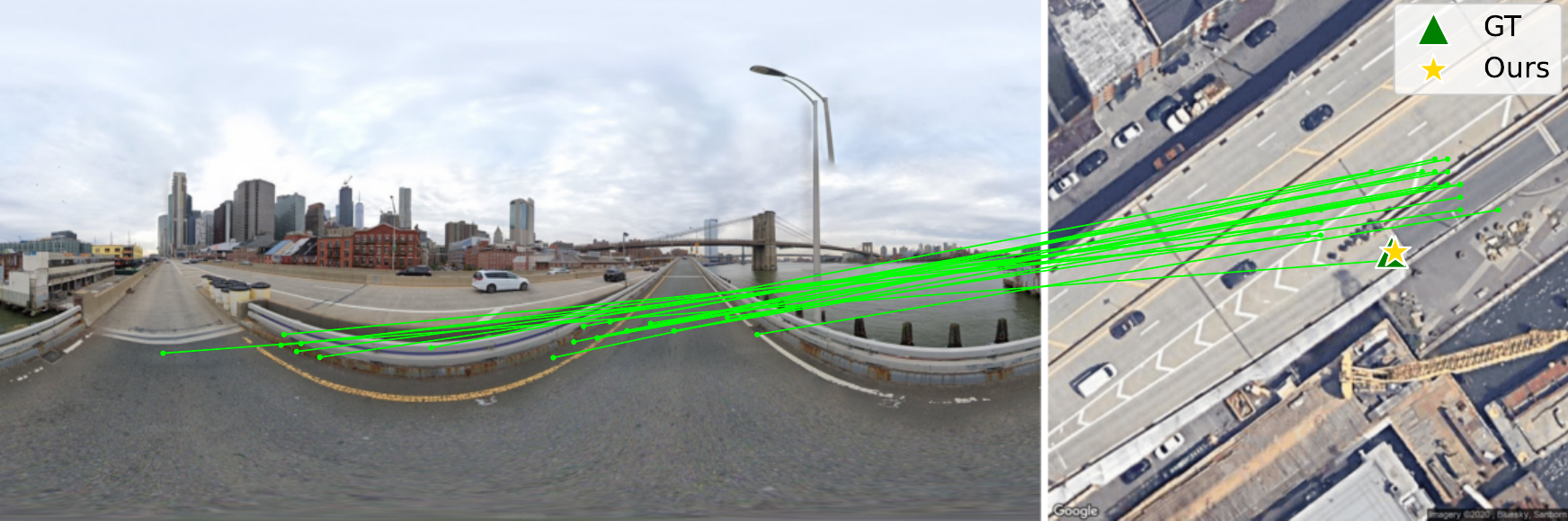}};
        \node[below right=2mm] at (a.north west) {(e)}; 
          }}
    \hfil
    \subfloat[]{%
    \tikz{\node (a) {\includegraphics{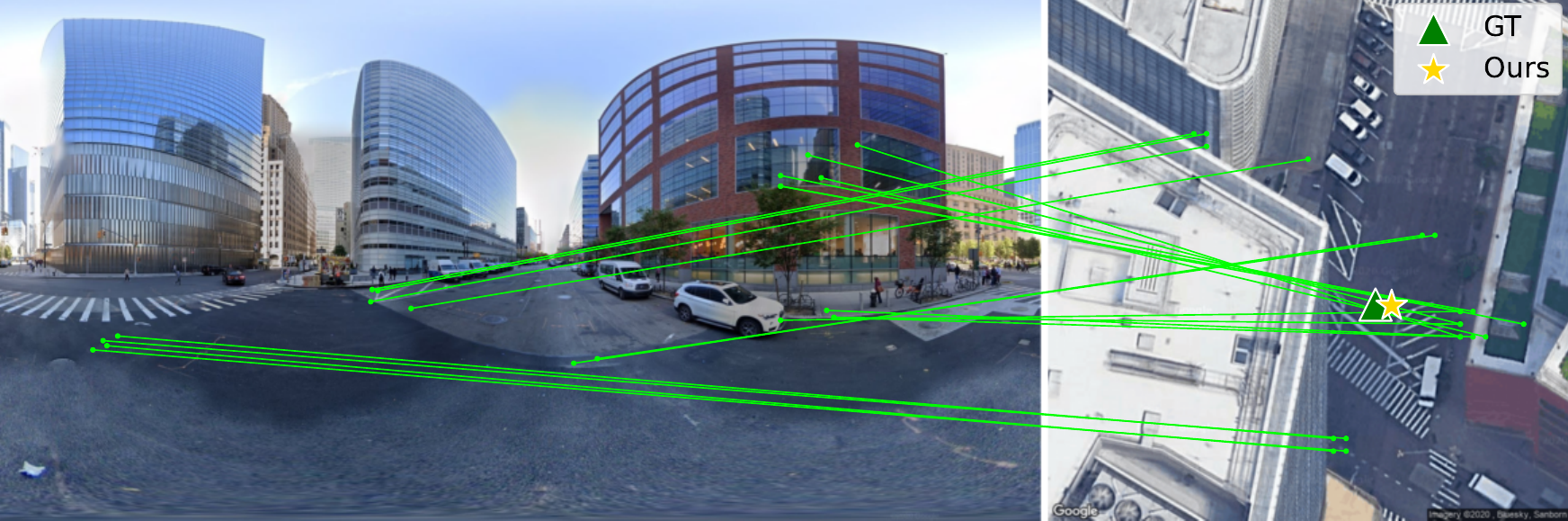}};
        \node[below right=2mm] at (a.north west) {(f)}; 
      }}
    \hfil
    \subfloat[]{%
    \tikz{\node (a) {\includegraphics{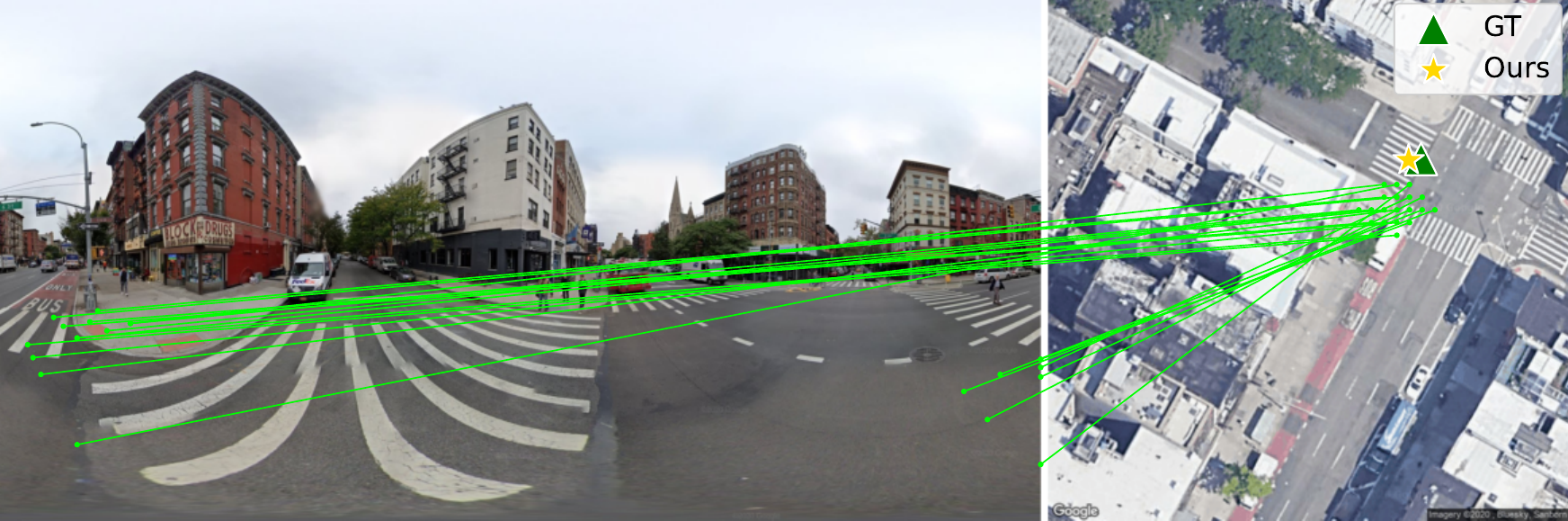}};
        \node[below right=2mm] at (a.north west) {(g)}; 
      }}
    \hfil
    \subfloat[]{%
    \tikz{\node (a) {\includegraphics{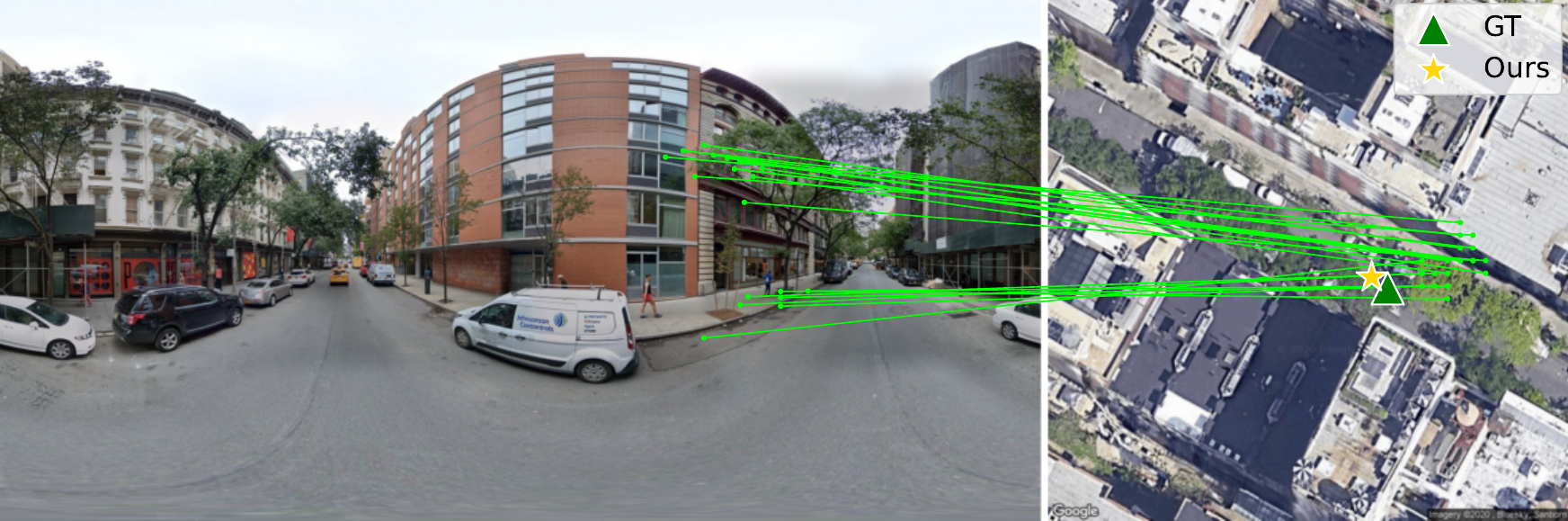}};
        \node[below right=2mm] at (a.north west) {(h)}; 
      }}
    \hfil
    \subfloat[]{%
    \tikz{\node (a) {\includegraphics{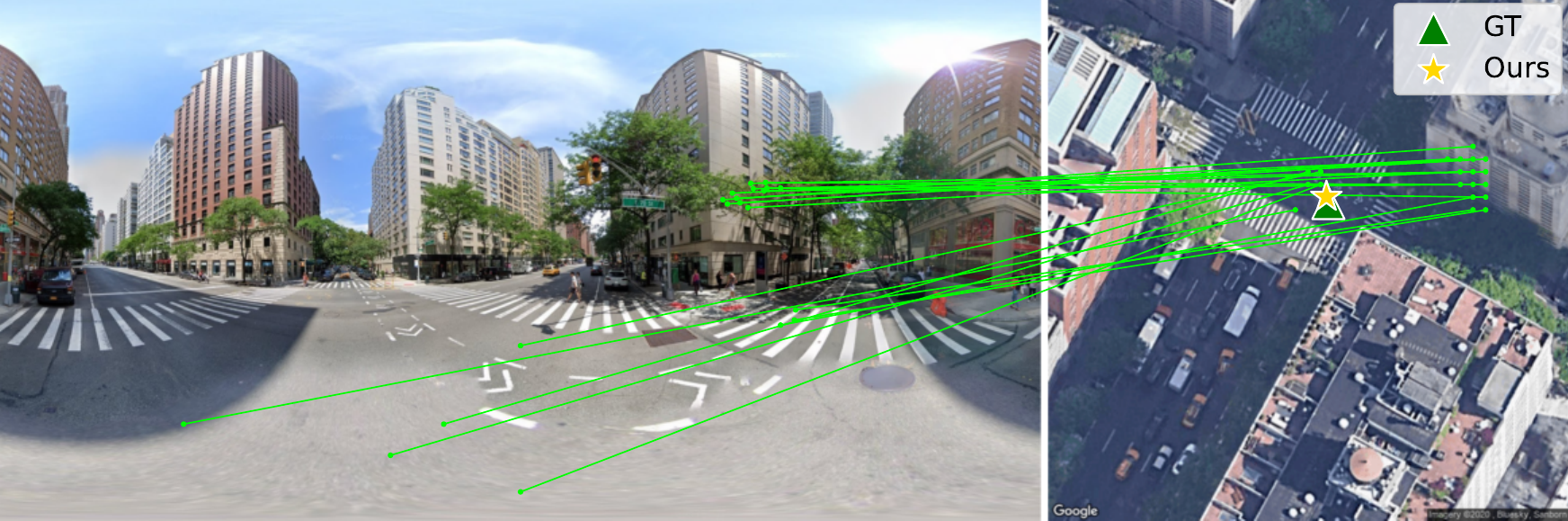}};
        \node[below right=2mm] at (a.north west) {(i)}; 
      }}
    \hfil
    \subfloat[]{%
    \tikz{\node (a) {\includegraphics{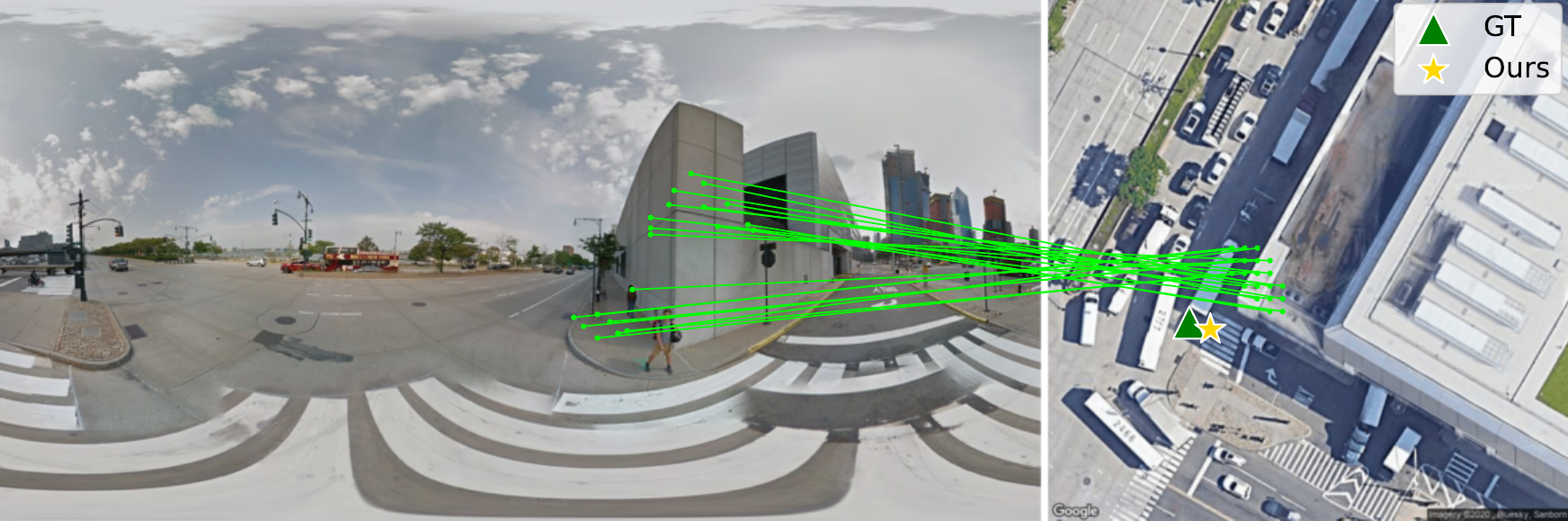}};
        \node[below right=2mm] at (a.north west) {(j)}; 
      }}
    \hfil
    \subfloat[]{%
    \tikz{\node (a) {\includegraphics{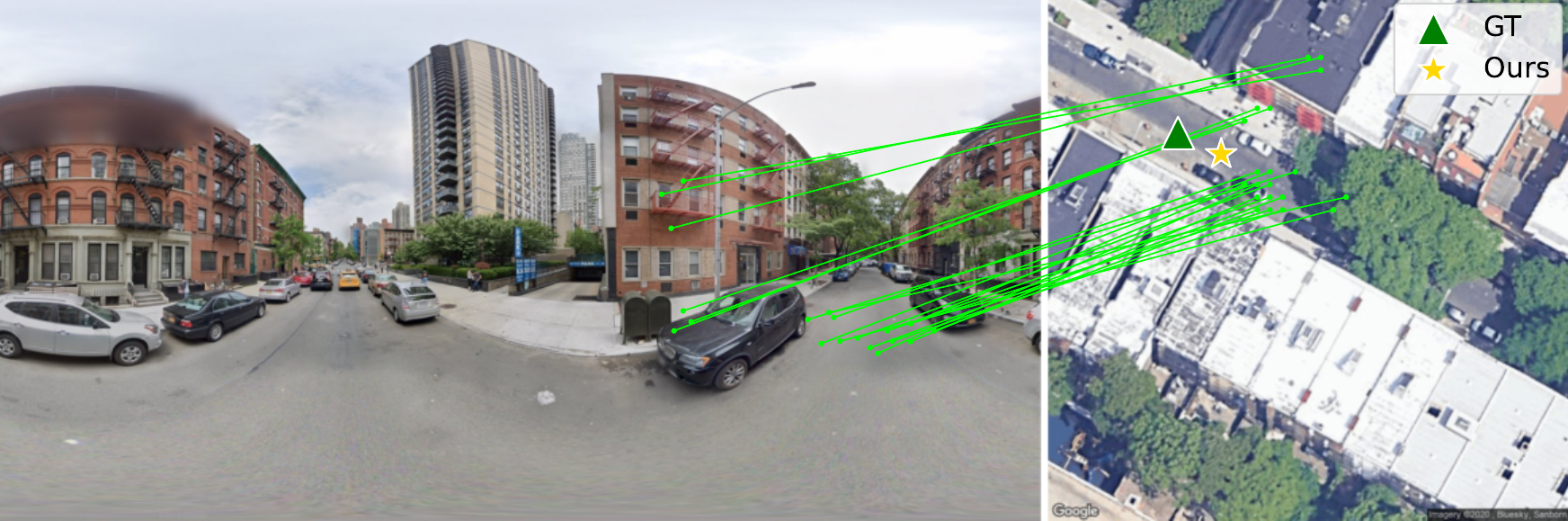}};
        \node[below right=2mm] at (a.north west) {(k)}; 
      }}
    \hfil
    \subfloat[]{%
    \tikz{\node (a) {\includegraphics{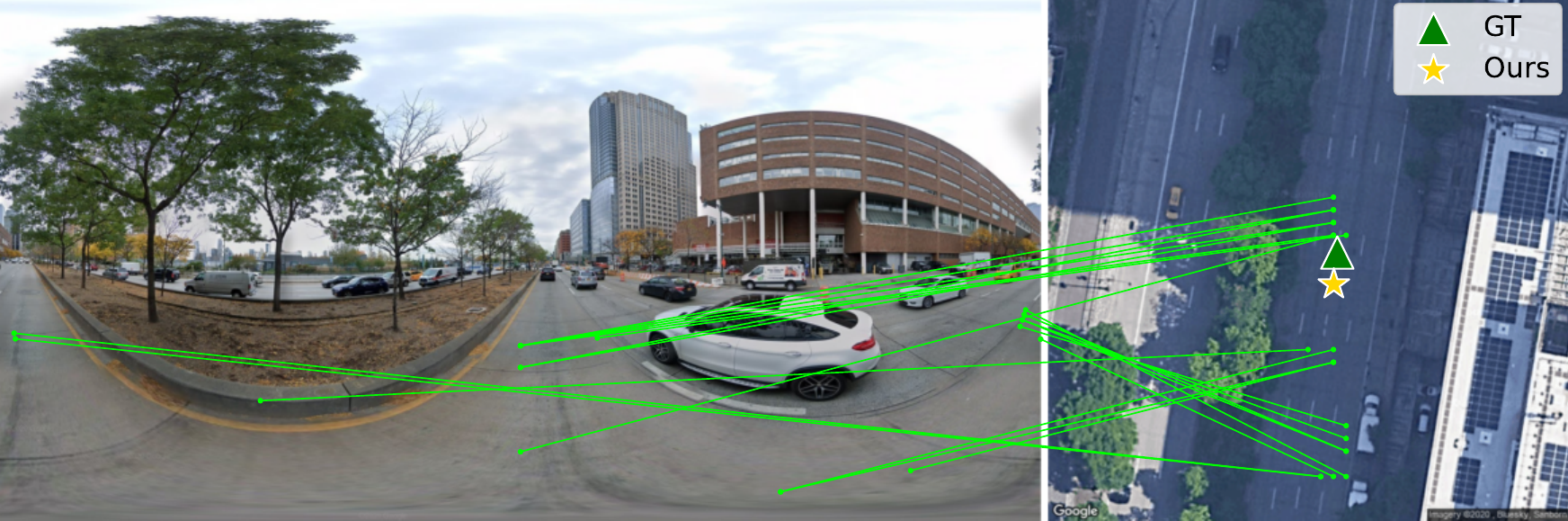}};
        \node[below right=2mm] at (a.north west) {(l)}; 
      }}
    \hfil
    \caption{Fine-grained feature matching results on VIGOR with known orientation. We show the 20 matches with the highest similarity scores. We find the 3D points using the selected height in the last pooling to BEV step and then project those points to the ground image.}
    \label{fig:extra_qualitative_results1}
\end{figure*}

\begin{figure*}[ht]
    \captionsetup[subfigure]{labelformat=empty}
    \tikzset{inner sep=0pt}
    \setkeys{Gin}{width=0.49\textwidth}
    \centering
    \subfloat[]{%
    \tikz{\node (a) {\includegraphics{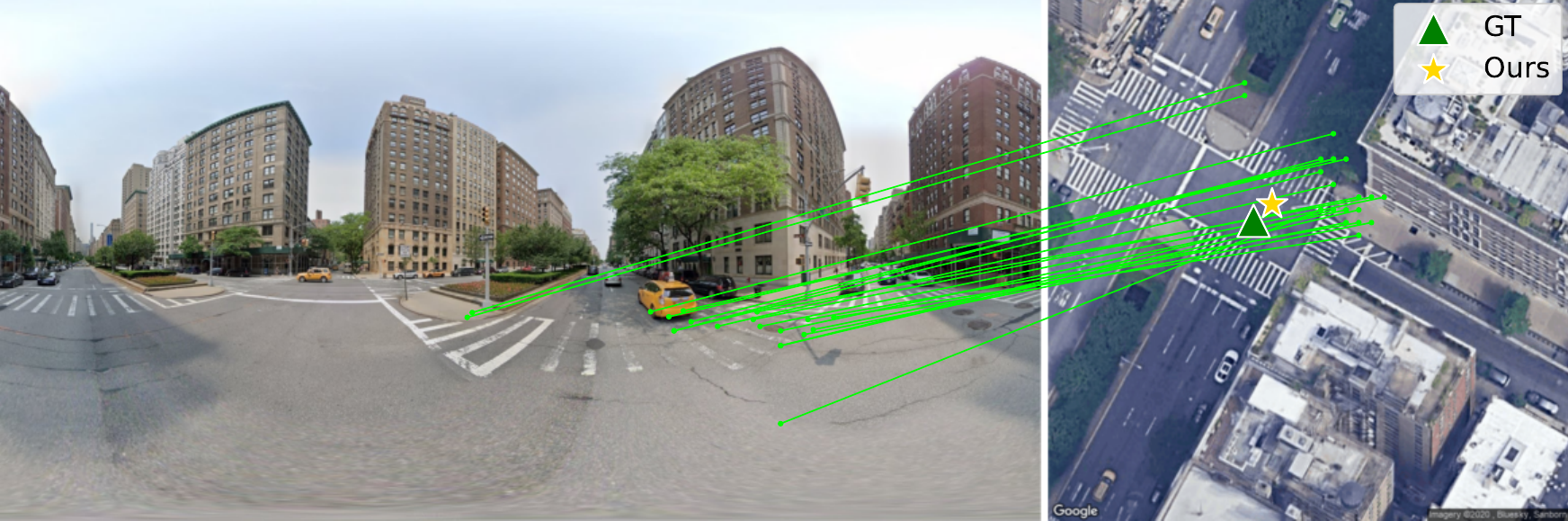}};
        \node[below right=2mm] at (a.north west) {(a)}; 
          }}
    \hfil
    \subfloat[]{%
    \tikz{\node (a) {\includegraphics{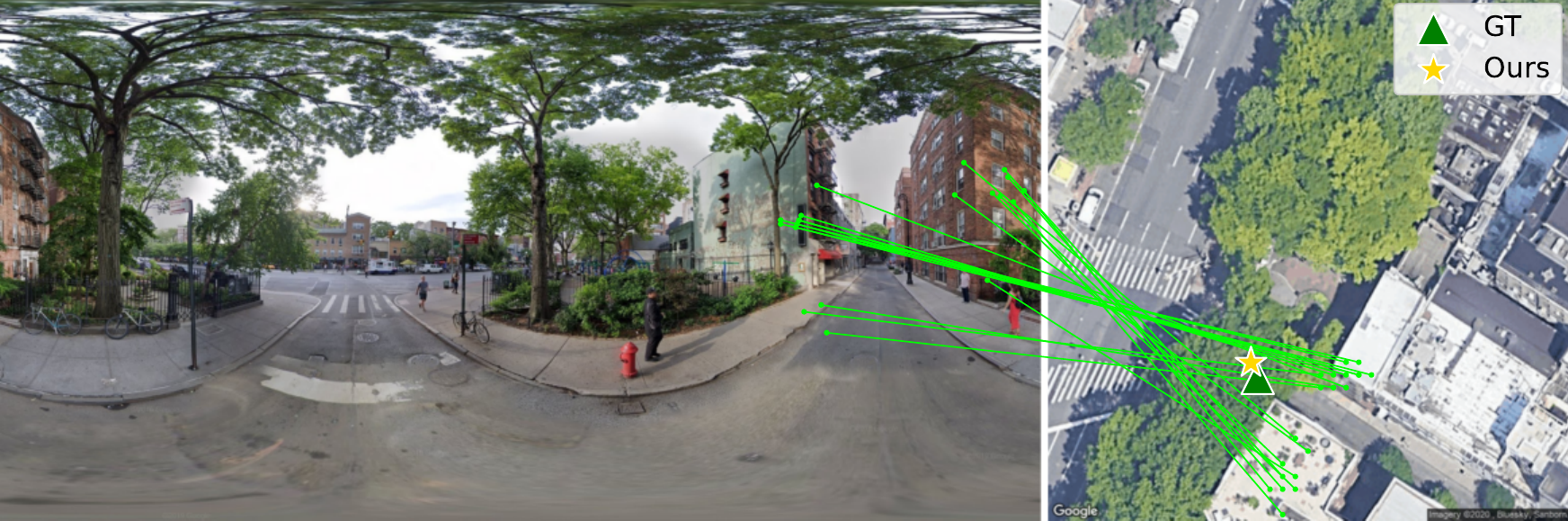}};
        \node[below right=2mm] at (a.north west) {(b)}; 
      }}
    \\
    \subfloat[]{%
    \tikz{\node (a) {\includegraphics{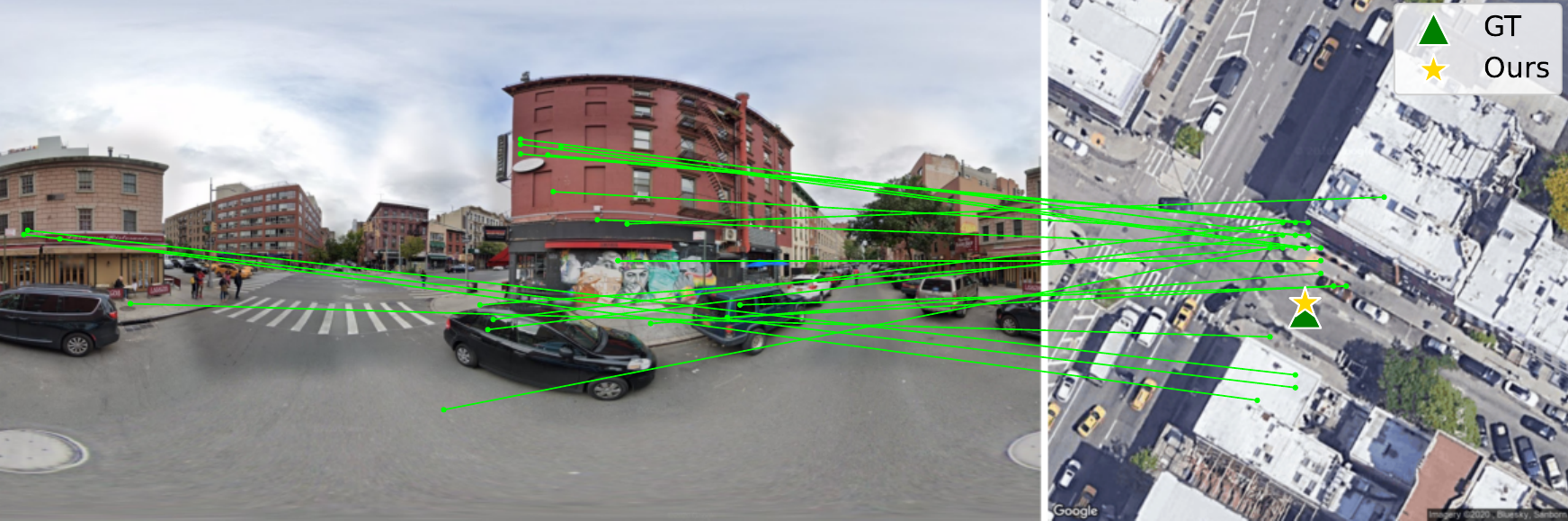}};
        \node[below right=2mm] at (a.north west) {(c)}; 
          }}
    \hfil
    \subfloat[]{%
    \tikz{\node (a) {\includegraphics{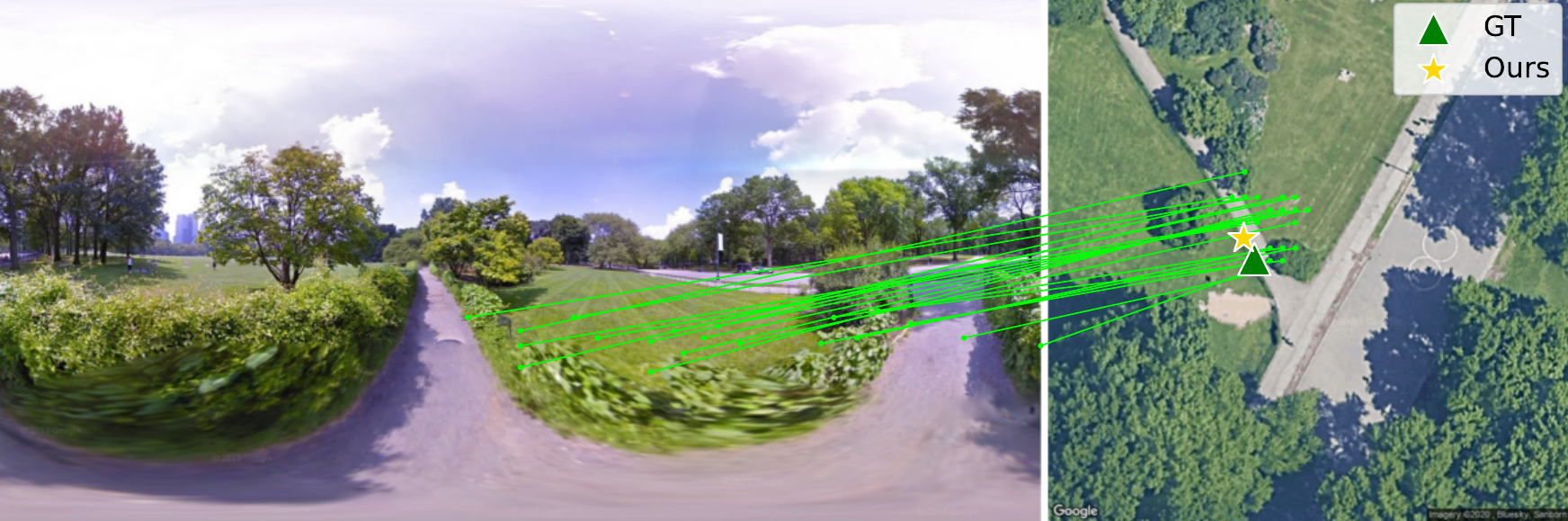}};
        \node[below right=2mm] at (a.north west) {(d)}; 
      }}
    \hfil
    \subfloat[]{%
    \tikz{\node (a) {\includegraphics{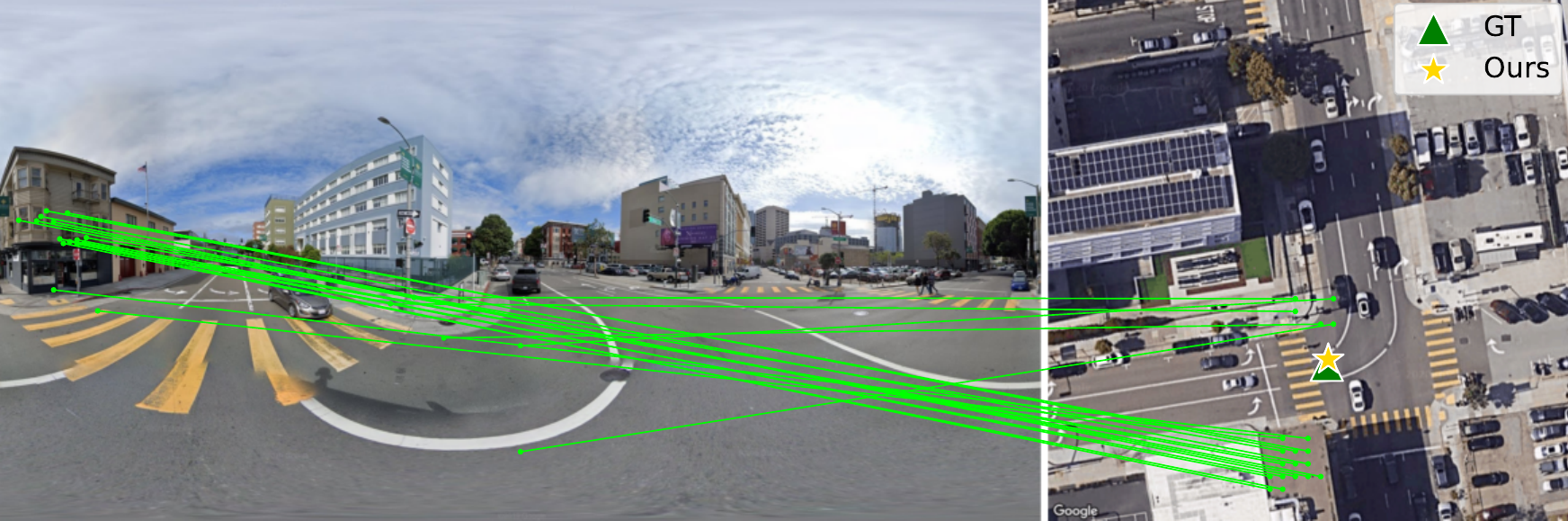}};
        \node[below right=2mm] at (a.north west) {(e)}; 
      }}
    \hfil
    \subfloat[]{%
    \tikz{\node (a) {\includegraphics{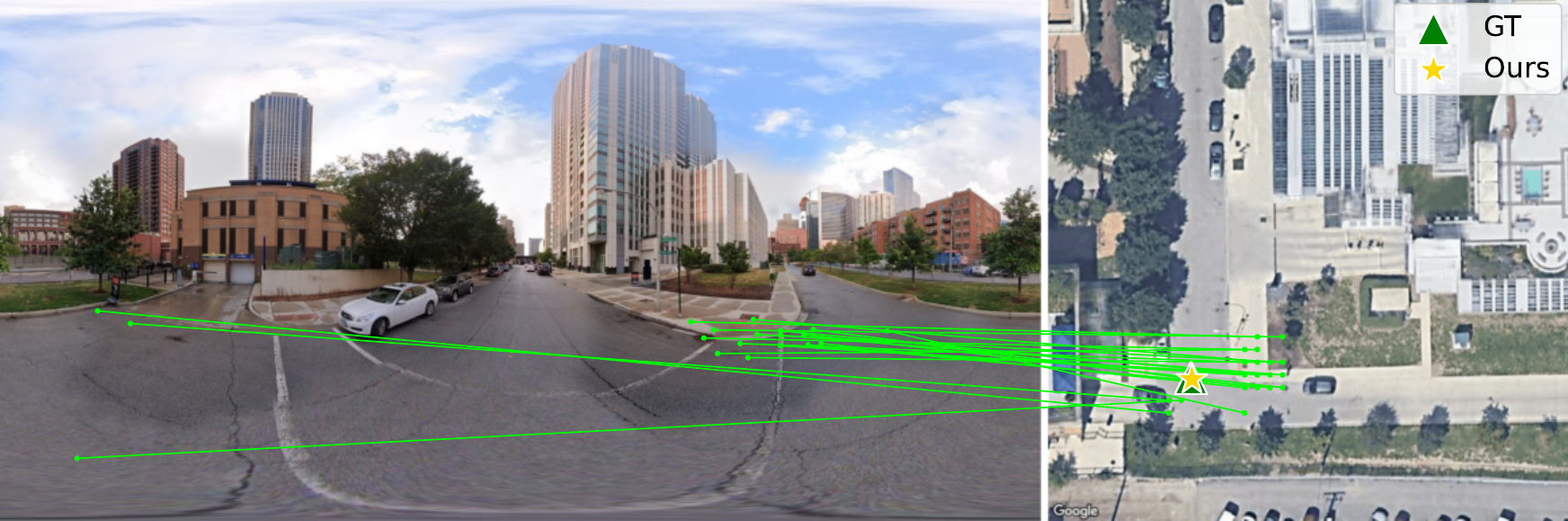}};
        \node[below right=2mm] at (a.north west) {(f)}; 
      }}
    \hfil
    \subfloat[]{%
    \tikz{\node (a) {\includegraphics{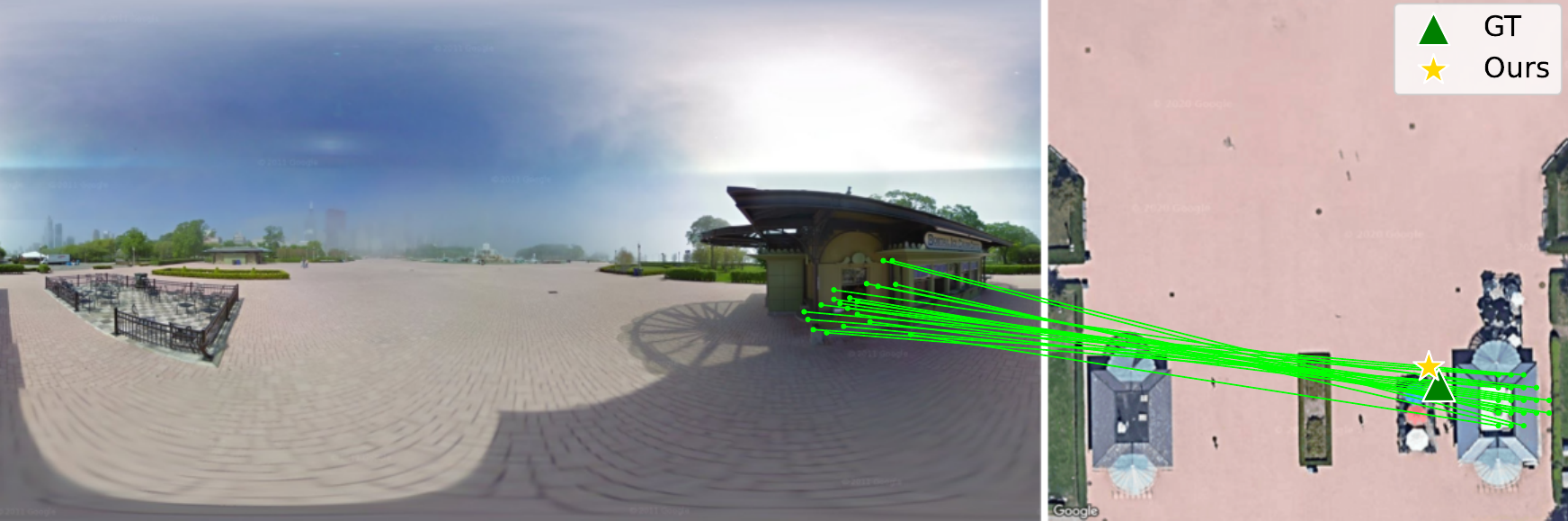}};
        \node[below right=2mm] at (a.north west) {(g)}; 
      }}
    \hfil
    \subfloat[]{%
    \tikz{\node (a) {\includegraphics{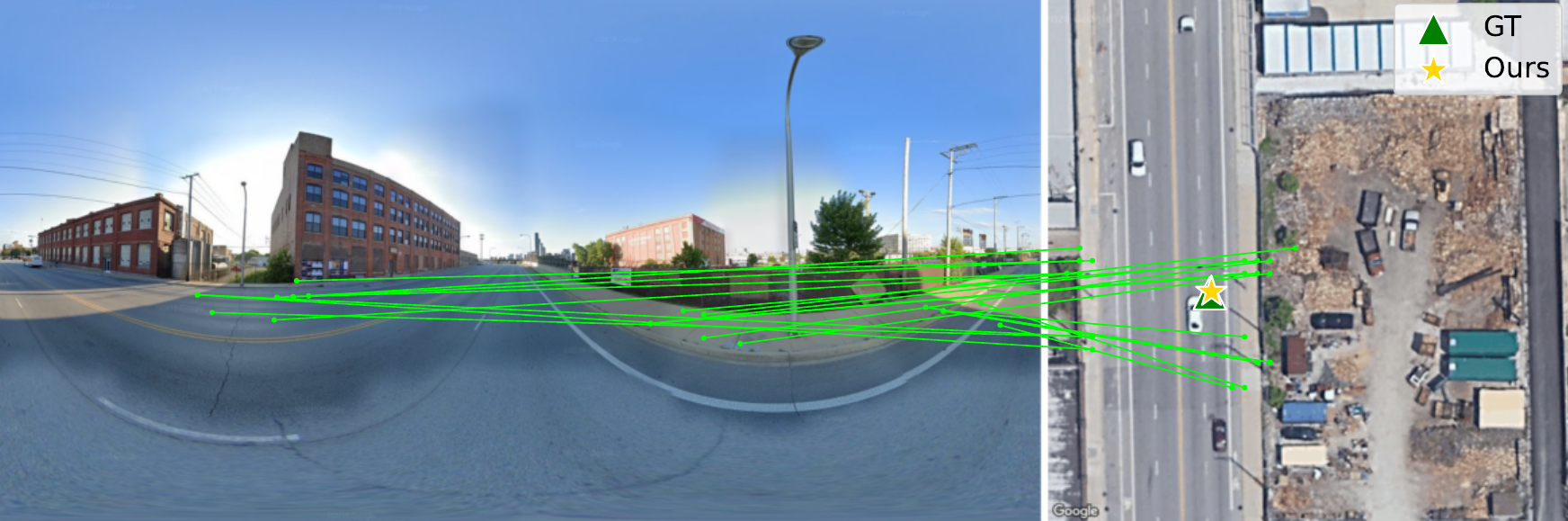}};
        \node[below right=2mm] at (a.north west) {(h)}; 
      }}
      \hfil
    \subfloat[]{%
    \tikz{\node (a) {\includegraphics{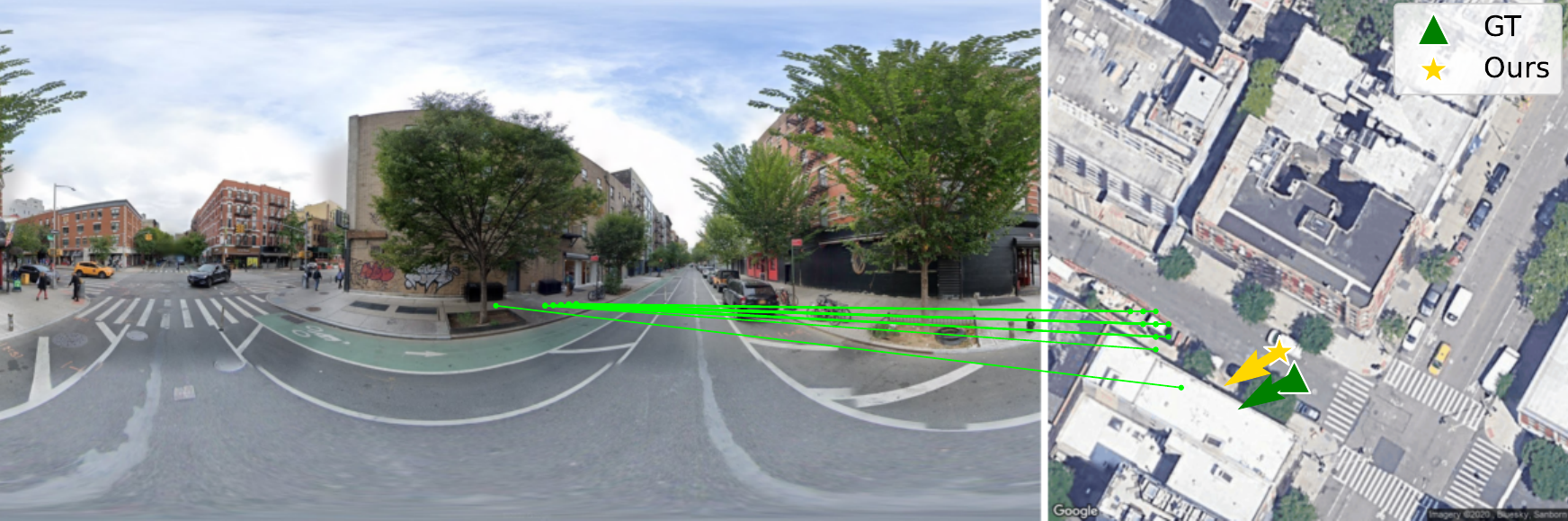}};
        \node[below right=2mm] at (a.north west) {(i)}; 
      }}
    \hfil
    \subfloat[]{%
    \tikz{\node (a) {\includegraphics{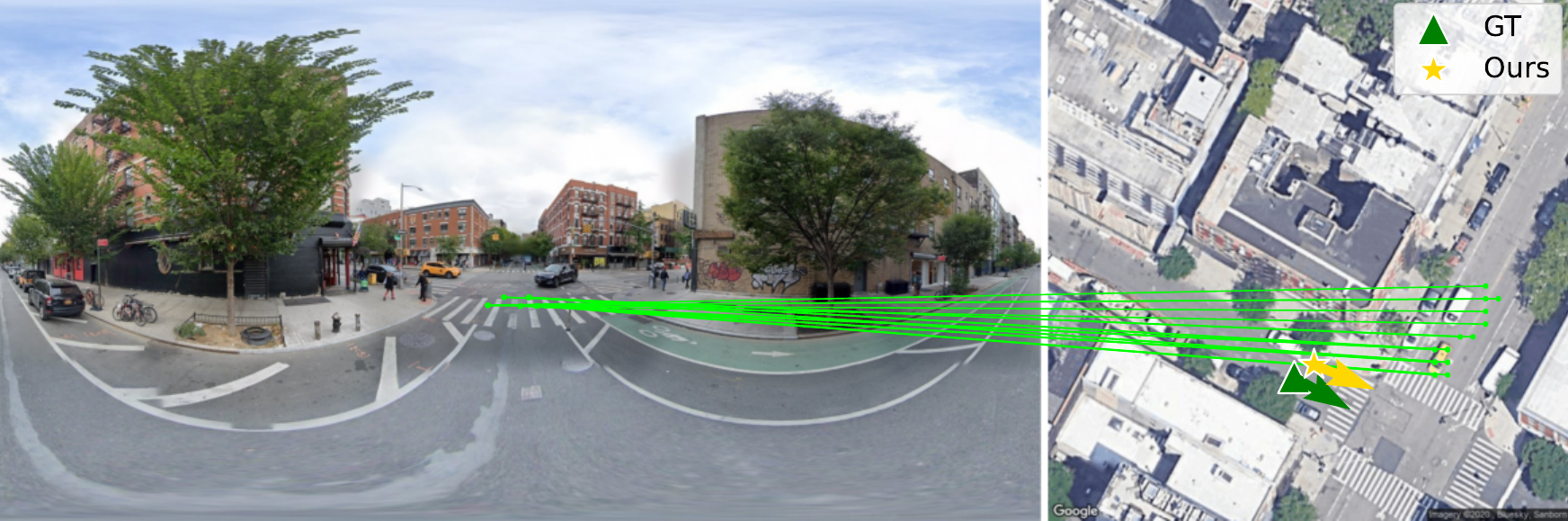}};
        \node[below right=2mm] at (a.north west) {(j)}; 
      }}
    \hfil
    \subfloat[]{%
    \tikz{\node (a) {\includegraphics{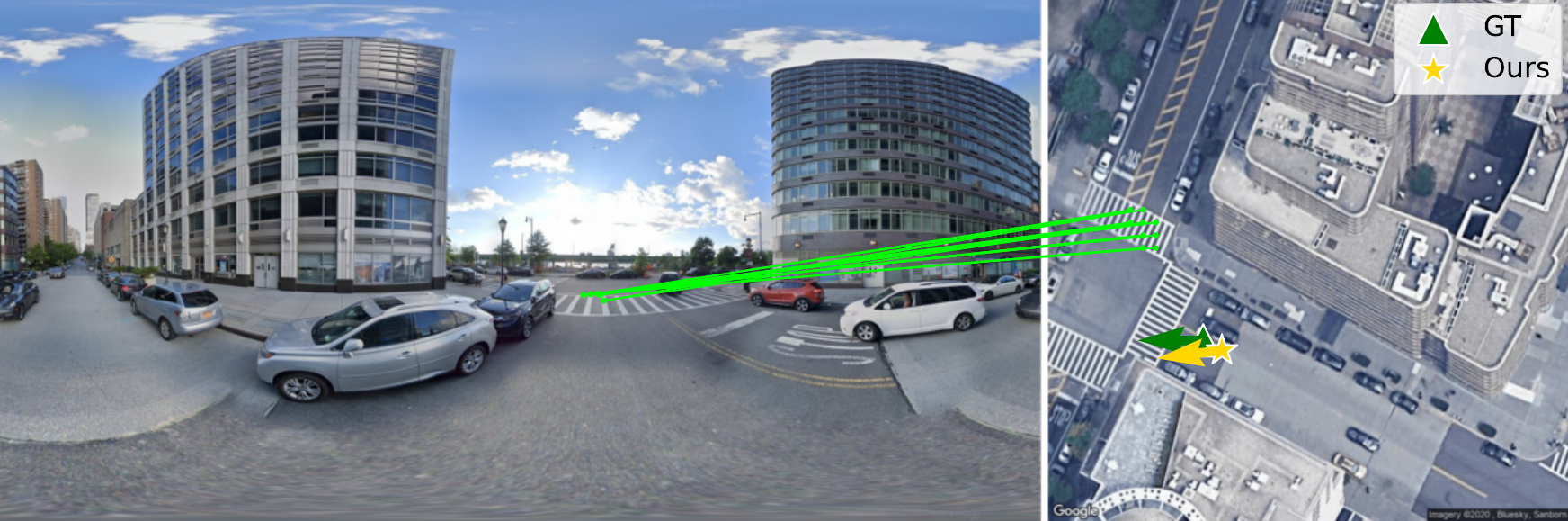}};
        \node[below right=2mm] at (a.north west) {(k)}; 
      }}
    \hfil
    \subfloat[]{%
    \tikz{\node (a) {\includegraphics{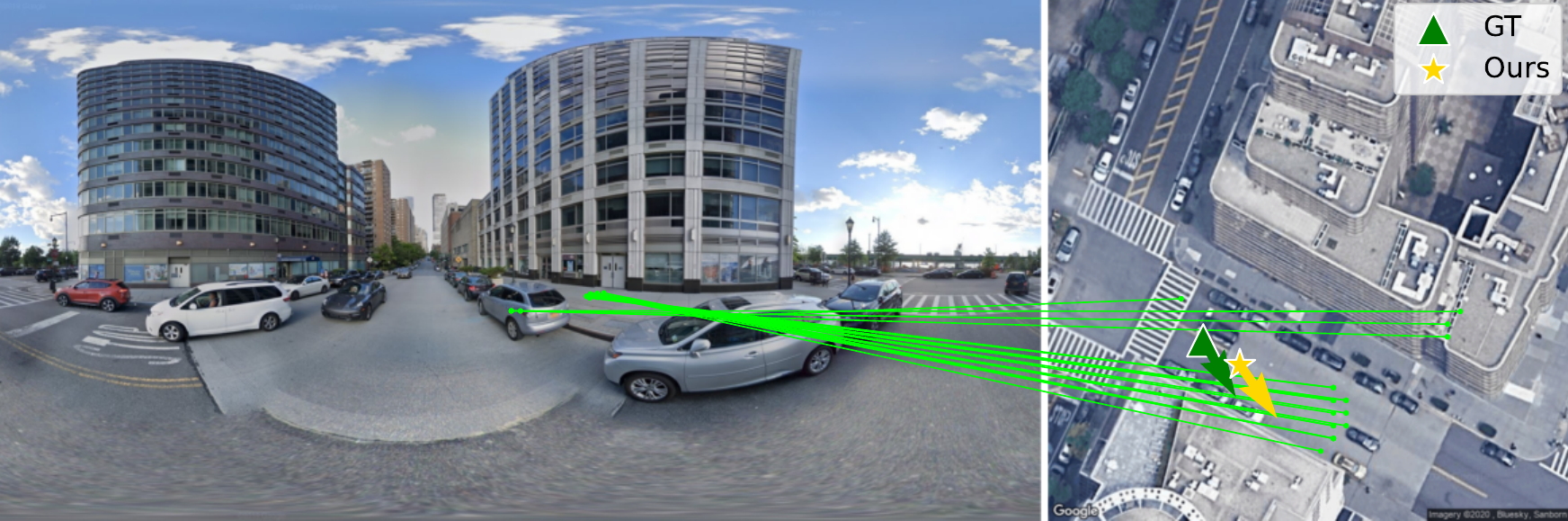}};
        \node[below right=2mm] at (a.north west) {(l)}; 
      }}
    \caption{Fine-grained feature matching results on VIGOR with known and unknown orientation. We show the 20 matches with the highest similarity scores. We find the 3D points using the selected height in the last pooling to BEV step and then project those points to the ground image.}
    \label{fig:extra_qualitative_results2}
\end{figure*}

Finally, in addition to the qualitative results presented in Sec.~4.4 of the main paper, we provide more examples in Fig.~\ref{fig:extra_qualitative_results1} and~\ref{fig:extra_qualitative_results2}.
In Fig.~\ref{fig:extra_qualitative_results1} (a)-(d), we include the same samples as in Fig.~\ref{fig:dino}, and our method produces reasonable matches.
In Fig.~\ref{fig:extra_qualitative_results2} (i)-(l), we show predictions with unknown orientation.

\newpage


\end{document}